\documentclass{article}

\usepackage{lmodern}
\usepackage[T1]{fontenc}

\usepackage[english]{babel}
\usepackage{microtype}
\usepackage{graphicx}
\usepackage[dvipsnames]{xcolor}
\usepackage{subcaption}
\usepackage{booktabs} %
\usepackage{multirow}
\usepackage{csquotes}
\usepackage{xurl}
\usepackage{natbib}
\usepackage{authblk}
\usepackage{enumitem}
\usepackage{colortbl}
\usepackage{listings}
\usepackage{fancyvrb}

\usepackage{booktabs}      %
\usepackage{tabularx}      %
\usepackage{threeparttable}%
\usepackage{ragged2e}      %
\usepackage{xcolor}        %

\newcolumntype{L}[1]{>{\RaggedRight\arraybackslash}p{#1}}
\newcolumntype{Y}{>{\RaggedRight\arraybackslash}X}

\usepackage{amsmath}
\usepackage{amsfonts}
\usepackage{amssymb}
\usepackage{mathtools}
\usepackage{amsthm}
\usepackage{listings}
\usepackage{fullpage}
\usepackage{parskip}
\usepackage{etoolbox}
\usepackage[breaklinks,colorlinks,bookmarks=False]{hyperref}
\usepackage[capitalize,noabbrev]{cleveref}

\ifundef{\abstract}{}{\patchcmd{\abstract}%
    {\quotation}{\quotation\noindent\ignorespaces}{}{}}

\hypersetup{
    urlcolor=TealBlue!40!black,
    citecolor=TealBlue!40!black,
    linkcolor=TealBlue!40!black,
    anchorcolor=TealBlue!40!black,
    citecolor=TealBlue!40!black,
    filecolor=TealBlue!40!black,
    menucolor=TealBlue!40!black,
    runcolor=TealBlue!40!black,
}

\theoremstyle{plain}

\theoremstyle{definition}

\theoremstyle{remark}

\author[,1]{Simon Frieder\thanks{Corresponding author: \href{mailto:simon.frieder@cs.ox.ac.uk}{\texttt{simon.frieder@cs.ox.ac.uk}}. 
}}
\author[2]{Jonas Bayer}
\author[4]{Sam Looi}
\author[2]{Jacob Loader}
\author[4]{Julius Berner}
\author[2]{Katherine M. Collins}
\author[1]{Andr\'{a}s Juh\'{a}sz}
\author[9]{Fabian Ruehle}
\author[7]{Sean Welleck}
\author[10]{Gabriel Poesia}
\author[11]{Ryan-Rhys Griffiths}
\author[2]{Adrian Weller}
\author[12]{Anirudh Goyal}
\author[3,1]{Thomas Lukasiewicz}
\author[13]{Cameron Freer}
\author[8]{Kevin Buzzard}
\author[2,6]{Timothy Gowers}

\affil[1]{{\normalsize University of Oxford}}
\affil[2]{{\normalsize University of Cambridge}}
\affil[3]{{\normalsize Vienna University of Technology}}
\affil[4]{{\normalsize Caltech}}
\affil[6]{{\normalsize Coll\`{e}ge de France}}
\affil[7]{{\normalsize Carnegie Mellon University}}
\affil[8]{{\normalsize Imperial College London}}
\affil[9]{{\normalsize Northeastern University}}
\affil[10]{{\normalsize Stanford University}}
\affil[11]{{\normalsize FutureHouse Inc.}}
\affil[12]{{\normalsize Meta}}
\affil[13]{{\normalsize MIT}}

\date{\vspace{-1em}}

\begin{document}

\title{Data for Mathematical Copilots: \\ Better Ways of Presenting Proofs for Machine Learning
}

\maketitle

\begin{abstract}
The datasets and benchmarks commonly used to train and evaluate the mathematical capabilities of AI-based mathematical copilots (primarily large language models) exhibit several shortcomings and misdirections. These range from a restricted scope of mathematical complexity to limited fidelity in capturing aspects beyond the final, written proof (e.g. motivating the proof, or representing the thought processes leading to a proof). These issues are compounded by a dynamic reminiscent of Goodhart's law: as benchmark performance becomes the primary target for model development, the benchmarks themselves become less reliable indicators of genuine mathematical capability.
We systematically explore these limitations and contend that enhancing the capabilities of large language models, or any forthcoming advancements in AI-based mathematical assistants (copilots or ``thought partners''), necessitates a course correction both in the design of mathematical datasets and the evaluation criteria of the models' mathematical ability. In particular, it is necessary for benchmarks to move beyond the existing result-based datasets that map theorem statements directly to proofs, and instead focus on datasets that translate the richer facets of mathematical research practice into data that LLMs can learn from. 
This includes benchmarks that supervise the proving process and the proof discovery process itself, and we advocate for mathematical dataset developers to consider the concept of \emph{``motivated proof''}, introduced by G. Pólya in 1949, which can serve as a blueprint for datasets that offer a better proof learning signal, alleviating some of the mentioned limitations.

\end{abstract}
\clearpage

\section{AI Systems for Mathematicians -- Present and Future}

There has been a rapid surge in research around AI for mathematics. For example, AlphaGeometry was shown to solve mathematical geometry problems on the level of the International Math Olympiad (IMO)~\citep{trinh2024solving}, Numina won the first AIMO Progress Prize,\footnote{\url{https://aimoprize.com/updates/2024-07-20-progress-prize-results}} by providing a model capable of turning mathematical questions into executable Python code, and previously GPT-4~\citep{openai2023gpt4} was shown to attain, in some cases, the performance of an undergraduate university student~\citep{frieder2023mathematical}. High accuracies on standard mathematics benchmarks~\citep{reid2024gemini} suggest that many benchmarks may be close to being solved.

\paragraph{Update since the December 2024 version.}
This manuscript was initially circulated in December 2024. Since then, reported performance on boxed-answer benchmarks has continued to increase, and some reasoning-focused systems report near-saturation on \textsc{MATH}-style evaluations~\citep{deepseek2025r1,qwen2024qwq32bpreview}. At the same time, many of the issues discussed in this paper have seen fewer changes than model scores: public math benchmarks remain largely static, scoring often still collapses rich outputs to a single correctness bit, and contamination risk persists as benchmarks diffuse into training corpora. The revisions in this version therefore focus on incorporating recent benchmark results while keeping the central claim unchanged: progress toward mathematical copilots depends on data and evaluation that represent intermediate reasoning and workflow competence, not only final answers.

The rapid progress and proliferation of ideas in this space -- while exciting and productive in many ways -- has led to a series of problems due to a lack of alignment across research directions and a lack of coherence around clear goals. 
Problems include: 
\begin{enumerate}
\item[(1)] some datasets being overstudied (e.g., the many versions of the GSM8K dataset~\citep{cobbe2021training}), whereas other data-related aspects (few datasets comprising advanced mathematics or reflecting tool use in mathematics) are neglected;
\item[(2)] various workflows and reasoning modes of how professional mathematicians conduct their research are not represented in datasets;
\item[(3)] different ways to express the same mathematical content (in formal language, vs.\ in natural language) have not been unified;
\item[(4)] scalability of evaluation that aligns with the goals of the tools being developed.
\end{enumerate}

These problems are not unique to AI for mathematics, but the broader machine learning community has begun systematic self-assessment efforts that the mathematical AI community has yet to undertake. For instance, \citet{bhardwaj2024curation} present an evaluation framework grounded in data curation principles from librarianship and archives, and use it to assess 60 datasets from the NeurIPS Datasets and Benchmarks track. Their analysis shows gaps in documentation around ethical considerations, data management, and environmental footprint, and they provide concrete recommendations for improving peer-review standards. No comparable audit exists for mathematical datasets. Such an effort would be valuable: mathematical datasets come with domain-specific concerns (e.g., difficulty calibration, formal-informal alignment, contamination risks) that generic ML dataset assessments do not address.

Despite these problems, the landscape of machine learning models that have been devised (and embraced by parts of the mathematical community) consists primarily of models that \textbf{lie between the two categories below}, based on the amount of interaction with humans they are designed for:

\begin{enumerate}
    \item \textbf{Highly specialized, \enquote{narrow} models} that fulfill all of the criteria of 1) being designed for a precisely defined mathematical domain or mathematical problem, 2) requiring domain knowledge to be used (such as formal languages as input), 3) that cannot explain themselves, and 4) do not allow free-form interaction. Examples of such specialized models vary widely in terms of how their foundation is set up (model architecture, training data), and in what they assist mathematicians. Specialized models could rely on symbolic reasoners, such as the recent solver for elementary geometry (Newclid~\citep{sicca2024newclid}, which supersedes AlphaGeometry~\citep{trinh2024solving}) or for inequality problems (\citep{wei2024proving}). Olympiad-level mathematics problems are within reach of both, and these assist mathematicians solely in solving closed-ended questions. Alternatively, such specialized models could simply be regression models, relying on raw mathematical data to infer new relationships between mathematical objects, such as for knot theory or representation theory~\citep{davies2021advancing} or group theory~\citep{he2023learning}, assisting mathematicians in finding conjectures (which then need to be proved by hand). Many further approaches exist~\citep{romera2024mathematical, fawzi2022discovering,hashemi2024can}, see Table~\ref{tab:narrow_models} for more information. These systems typically require a significant amount of specialization by a user, such as mastering a formal input language (e.g., AlphaGeometry and Newclid, which both require mastery of the JGEX formal language) or understanding the machine learning model itself (e.g.,~\citep{davies2021advancing}) beyond the necessary mathematical domain knowledge. The data on which they operate also varies strongly: Symbolic solvers rely on a symbolic language; models like~\citep{davies2021advancing} rely on tabular data, while models like~\citep{hashemi2024can} combine different data types.

\begin{table}[t]
\centering
\footnotesize
\setlength{\tabcolsep}{5pt}
\renewcommand{\arraystretch}{1.12}

\begin{threeparttable}
\begin{tabularx}{\linewidth}{@{} L{3.25cm} L{2.35cm} L{2.75cm} Y L{1.7cm} @{}}
\toprule
\textbf{System} &
\textbf{Reference} &
\textbf{Architecture class} &
\textbf{How it helps mathematicians} &
\textbf{User understanding} \\
\midrule

\multicolumn{5}{@{}l}{\textit{Closed-ended problem solving (verifiable / near-verifiable)}}\\[-1pt]
\addlinespace[2pt]
AlphaGeometry &
\citet{trinh2024solving} &
Hybrid (neural + symbolic) &
Solves Euclidean geometry problems via learned components plus symbolic deduction; uses a domain-specific formal encoding (JGEX). &
IF \\
Newclid &
\citet{sicca2024newclid} &
Symbolic (search / ATP-style) &
Solves Euclidean geometry problems in JGEX, producing checkable derivations. &
IF \\
Inequality prover &
\citet{wei2024proving} &
Symbolic (automation) &
Automates structured inequality proving in restricted families, producing derivations that can often be mechanically verified.
&
IF \\

\addlinespace[4pt]
\multicolumn{5}{@{}l}{\textit{Conjecture, counterexample, and construction discovery (requires downstream validation)}}\\[-1pt]
\addlinespace[2pt]
Graph-building RL counterexamples &
\citet{Wagner:2021Constructing} &
RL policy (often MLP/GNN) &
Finds counterexamples by constructing discrete objects in an environment with an automatically scored objective. &
PL \\
PatternBoost &
\citet{Charton:2024Patternboost} &
Transformer + local search &
Discovers combinatorial constructions by alternating neural generation with search, targeting automatically checkable properties. &
PL \\
Invariant-correlation models (e.g., knot theory) &
\citet{davies2021advancing} &
Supervised regression (often linear / tree-based) &
Suggests conjectural relationships between computable invariants; outputs hypotheses that mathematicians then prove or refute. &
OB \\
Structured/tabular ML in group theory &
\citet{he2023learning} &
Supervised learning (classifier/regressor) &
Produces conjectures or predictions from engineered representations of algebraic objects; requires mathematical validation. &
OB \\

\addlinespace[4pt]
\multicolumn{5}{@{}l}{\textit{Formal proof engineering (assistance within an ITP ecosystem)}}\\[-1pt]
\addlinespace[2pt]
ImProver (Lean) &
\citet{ahuja2024improver} &
Transformer (LLM program synthesis) &
Edits or refactors Lean proofs under constraints (e.g., length), helping with proof maintenance and presentation. &
IF \\

\bottomrule
\end{tabularx}

\begin{tablenotes}[para,flushleft]
\footnotesize
\item\textbf{User understanding}: \textbf{BB} = black-box use; \textbf{IF} = interface or formalism (e.g., Lean, JGEX);
\textbf{PL} = pipeline configuration and objective (representation, search, scoring);
\textbf{OB} = feature or representation choices and the validation procedure used to interpret outputs.

\end{tablenotes}
\end{threeparttable}

\caption{Highly specialized (“narrow”) systems cited in this paper, summarized by architecture class, primary mathematical role, and the typical user understanding required.}
\label{tab:narrow_models}
\end{table}

    \item \textbf{General purpose, \enquote{broad} models} that can interact with humans via natural language, understand input images, and use tools (being able to call, in particular, specialized models, which can be viewed as a particular kind of \enquote{tool}, as we argue below), can provide general assistance to a mathematician (both with routine tasks, such as literature search, and domain-dependent tasks, such as counterexample search) and are not tied to a specific area of mathematics. These are what we refer to as \textbf{\enquote{Mathematical Copilots}}, as their goal is not simply to provide correct proofs but also to enhance the user's understanding of the results and guide them towards their own discoveries~\citep{collins2023evaluating}. Currently, the best candidates for these systems are large language models (LLMs), which we take to include multi-modal models that also admit images as components of the prompt. We contend that LLMs represent early predecessors of future general-purpose models (mathematical copilots), ultimately acting as mathematicians' \enquote{thought partners}~\citep{collins2024building}. Anecdotal evidence of such usefulness of current LLMs has been presented in various discussion forums,\footnote{\url{https://www.reddit.com/r/math/comments/14p6j5c/terence_tao_on_using_gpt4_to_help_with_math/}} while noting the failures that LLMs often succumb to. However, there are concrete ways in which LLMs can be improved to become mathematical copilots: As indicated by the tool-integrated-reasoning approach, championed by models such as ToRA~\citep{gou2023tora} or Numina,\footnote{\url{https://huggingface.co/AI-MO/NuminaMath-7B-TIR}} general LLMs can incorporate other symbolic systems to which they can delegate symbolic tasks to. The mixture-of-experts paradigm~\citep{cai2024survey}, on the other hand, highlights different ways in which LLMs can incorporate other LLMs as specially trained \enquote{modules} to which they can delegate mathematical tasks that are not symbolic in nature but still require particular mathematical skills. Such a hierarchical system can help mathematicians both on routine academic tasks (see~\citep{frieder2023large} for a short, high-level overview) and specialized ones. 
    We envision that such models will require significantly less effort from users to use them, compared to specialized models, being self-explanatory, where natural language is the main mode of interaction.
    (While the outlined architectural approaches provide a viable step forward, in some cases, no data exists on which to train these models; see below.)
\end{enumerate}   
In addition, a third class of models will likely be developed:
\begin{enumerate}
    \setcounter{enumi}{2}
    \item \textbf{Fully automated, \enquote{universal} models} that can generate mathematical theorems and theories autonomously, merely by being directed to a mathematical result. We envision that these systems will be the true successors of automated theorem provers (ATPs) such as the Vampire~\citep{riazanov2002design, kovacs2013first} or E~\citep{schulz2002brainiac} ATPs. These systems have not been widely adopted by the mathematical community. While there have been attempts to imbue these systems with (non-LLM) machine learning techniques~\citep {holden2021ML4ATP}, these have not yet dramatically increased their performance. Nonetheless, once general-purpose models have been established, it is conceivable one could put these in automation loops to obtain \enquote{AI mathematicians}. For the domain of AI, this has already been recently investigated in the form of an \enquote{AI Scientist}~\citep{lu2024ai} that autonomously generated machine learning research articles. These are systems that conceivably will require minimal human input, merely guiding them towards certain theorems that one wants to see analyzed (proved, refuted, with comments on whether slightly different variations of a theorem, with slightly altered hypotheses, admit a proof). 
\end{enumerate} 

We focus in this article on \textbf{general purpose models}, as outlined above, which are \textbf{mathematical copilots}. LLMs are currently the systems that hold the most promise to become, given a better data foundation, general purpose models. Fully automated, universal models are currently out of reach, as they are contingent on the former category of models. These are the systems that are ultimately sought when one speaks of \enquote{AI Mathematicians}~\citep{bengio2024machine}. However, without the proper datasets, it is unlikely that these will come into existence. Hence, our focus is on the \textbf{data that is needed to arrive at general-purpose models, the mathematical copilots}.
We have deliberately avoided drawing a hard boundary among the properties a model must fulfill to be placed in one of these three because the boundaries will be fluid, as the levels of automation and required human interaction are continuous scales. %

Currently, the existing datasets only support highly specialized models and do not put LLMs on a trajectory to achieve general-purpose models. In this article, we highlight the limitations of current datasets (Section~\ref{sec: past}, which is retrospective) and necessary changes in terms of data that will be needed to train the next generation of LLMs that approximate what we outlined general-purpose models to be (Section~\ref{sec: future}, which is prospective).

Several general principles have emerged from decades of machine learning research that shape expectations about AI progress. Perhaps the most influential is Sutton's ``bitter lesson''~\citep{sutton2019bitter}, which observes that general methods leveraging computation (particularly search and learning) consistently outperform approaches that encode human domain knowledge. More recently, researchers have argued that open-endedness -- the capacity for systems to continually generate novel, learnable artifacts -- is essential for artificial superhuman intelligence~\citep{hughes2024openendedness}. However, there are reasons to believe these principles may not straightforwardly transfer to mathematical AI. Mathematics demands logically rigorous proofs that must be verifiable, not merely plausible; pattern matching and statistical regularities, which suffice in domains like vision or game-playing, are insufficient for mathematical reasoning. Moreover, the ``bitter lesson'' presupposes abundant data that scales with computation, yet high-quality mathematical data, particularly at research level, is scarce, and much mathematical knowledge is implicit in workflows and heuristics that textbooks do not record. Open-endedness in mathematics requires more than generating novel outputs: it requires outputs that are \emph{correct} and \emph{meaningful} within a precise logical framework. These considerations suggest that progress in AI for mathematics may require domain-specific innovations in data, evaluation, and methodology that go beyond what has succeeded elsewhere in machine learning. 

\paragraph{The limits of end-to-end generation for mathematics.}
A recurring theme in machine learning is that end-to-end approaches -- where models learn directly from inputs to outputs without explicit intermediate structure -- often outperform systems with hand-designed pipelines~\citep{huang2023mlagentbench}.
However, we argue that for mathematical reasoning, pure end-to-end generation must be abandoned.
Mathematics demands proofs that are not merely plausible but logically valid, and errors in intermediate steps propagate to invalidate entire arguments.
Recent reasoning models address this by incorporating explicit chain-of-thought processes and, in formal settings, by interacting with proof assistants that verify each step~\citep{first2023baldur, song2024towards}.
The success of these hybrid approaches, which combine neural generation with symbolic verification, suggests that mathematical AI requires structured reasoning modes rather than monolithic sequence-to-sequence generation.

Our main reader audience consists of machine learning researchers, as we intend this article to raise awareness of these issues. Automated theorem provers (ATPs) and interactive theorem provers (ITPs), which have a decades-old history~\citep{harrison2014history}, also had the ambition of realizing the goal of implementing mathematical copilots. Unfortunately, history has shown that if the computer science community and the mathematical community do not operate in lockstep, there is a risk of obtaining systems that will not be of interest to practicing research mathematicians; maybe controversially,~\citet{blanchette2012more} mentions that certain efforts to make these systems more user-friendly have perhaps reached a \enquote{plateau}. (Nonetheless, ATPs and ITPs have, in other domains, contributed significantly to driving the field of software automation forward.) 

One of the goals of this paper is to make sure that the renewed effort by computer scientists and machine learning researchers to provide mathematicians with tools that make their trade easier this time is on a track to converge to the needs of mathematicians. We hope that one of the first manifestations of this will be in the form of datasets that better capture the various rich facets of mathematical research so LLM creators can provide scores on such benchmarks that are more informative for mathematicians regarding how useful the LLM is for actual daily mathematical practice. While the most-used datasets, GSM8K and MATH, are indicative of LLMs' reasoning abilities, they are largely irrelevant to mathematical practice.

\section{An Overview of Data-Related Issues}

Without using tools specialized for mathematics, the current general models, LLMs, simply by well-designed training methodologies and trained on mathematical data, have demonstrated unprecedented capabilities in generating humanlike mathematical text, solving complex problems, and even engaging in creative problem-solving. For example, GPT-4 has performed promisingly on undergraduate-level mathematics formulated in natural language, as pointed out by~\citet{frieder2023mathematical}. Math-Specialized Gemini 1.5 Pro~\citep{reid2024gemini}, a commercial model by Google not available to the public, has been reported to have an accuracy of over 90\% on the MATH dataset~\citep{hendrycks2021measuring}, one of the most widely-used datasets for testing mathematical ability.  This score has recently been replicated by an open-weight model, QwQ.\footnote{\url{https://qwenlm.github.io/blog/qwq-32b-preview/}} Attaining such a high score is equivalent to achieving the ability of an IMO gold medallist (according to~\citep{hendrycks2021measuring}). However, to date, no reports have been made where the mathematical benchmarks of QwQ are tested for contamination; not undertaking tests may lead to this score not being reproducible on a different dataset from the same difficulty and problem type distribution~\citep{yang2023rethinking, xu2024benchmarking}.

\paragraph{Benchmark saturation on \textsc{MATH}.}
Recent reasoning models approach or exceed 90\% accuracy on the \textsc{MATH} benchmark~\citep{hendrycks2021measuring}, which was considered extremely challenging at introduction (models scored below 7\%).
On \textsc{MATH}-500, a commonly used 500-problem subset, DeepSeek-R1 reports 97.3\%~\cite{deepseek2025r1} and QwQ-32B-Preview reports 90.6\%~\cite{qwen2024qwq32bpreview}.
Earlier models scored lower: Gemini 1.5 Pro achieved 77.9\% on full \textsc{MATH} with majority voting~\cite{reid2024gemini}.
This rapid progression leaves limited headroom under boxed-answer scoring and motivates harder benchmarks like FrontierMath~\citep{glazer2024frontiermath}.

DeepTheorem reports a large corpus of informal IMO-style theorems paired with proofs, and uses entailed and contradictory variants to provide a binary reward signal for reinforcement learning in informal theorem proving~\citep{deeptheorem2025}. Notably, this is a step beyond standard boxed-answer scoring: the supervision and evaluation are tied to logical consistency across theorem variants rather than only matching a numeric final answer.

These benchmark improvements do not, by themselves, resolve the evaluation and data problems that matter for mathematical copilots. First, high accuracy under boxed-answer scoring (still common in public benchmarks) leaves ambiguity about what changed, since multiple failure modes in the reasoning trace can be invisible once only the final answer is checked. Second, evidence from out-of-distribution tests suggests that strong benchmark scores can coexist with brittle reasoning on simple variants of problems~\citep{nezhurina2024alice}. Third, as scores rise, contamination and memorization become harder to rule out with simple string-based decontamination, especially when mathematically equivalent problems can be phrased in many surface forms~\citep{xu2024benchmarking}. These points motivate evaluation sets with stronger verification and a shift toward datasets that expose intermediate structure.

Further notable general-purpose LLMs are open-weight models with strong reported performance on baseline datasets, such as MATH and GSM8K~\citep{cobbe2021training}, the DeepSeek family of models~\citep{shao2024deepseekmath, liu2024deepseek, zhu2024deepseek, xin2024deepseek}, and the Qwen family of models~\citep{bai2023qwen, yang2024qwen2, hui2024qwen2, yang2024qwen2math}. Other LLMs focus more on specific abilities such as MathPrompter~\citep{imani2023mathprompter}, which associates a confidence value to arithmetic problem solutions, MathVista~\citep{lu2023mathvista}, which solves geometric reasoning problems, WizardMath~\citep{luo2023wizardmath}, or Llemma~\citep{azerbayev2023llemma}, which was specialized through further pretraining on a math corpus, and is both open-weight and open-source.
A special use case where LLMs have found success is in interactive theorem provers (ITPs) such as Isabelle~\citep{nipkow2002isabelle} or Lean~\citep{moura2021lean4}, where they are used to prove theorems most prominently in the context of generating proofs of given formal statements~\citep{first2023baldur,zheng2023lyra,xin2023lego}, 
autoformalization~\citep{szegedy2020promising,jiang2022draft}, or providing code snippets of formal mathematics~\citep{song2024towards}. For a survey of deep learning approaches on both formal and informal (natural language) mathematics, we refer to~\citet{lu2022survey}, and for more recent information on language models, including their performance on mainstream datasets, we refer to~\citet{zhao2023survey}. 

This paper argues that the current dataset landscape does not support the advancement of such general-purpose AI systems, in particular LLMs, towards a level of mathematical performance that makes them usable as daily \emph{mathematical thought partners}~\citep{collins2024building} that help mathematicians push the boundaries of what is known, and capture the richness of mathematical reasoning and invention that human mathematical minds are capable of~\citep{zhang2023cogsciMath, dehaene2011number, feigenson2004core}. Our focus thus includes, in particular, research-level mathematics, which involves many steps beyond deriving or formalizing proofs -- the aspects of mathematical practice that are currently most strongly represented in data. 

The spectrum of tasks a copilot can assist with is not solely related to mathematics: programmers presently carry out programming tasks, such as fixing bugs or other GitHub issues~\citep{jimenez2024swebench}, using various code-generating copilots, such as the GitHub Copilot~\citep{chen2021evaluating}; the emerging Lean Copilot ecosystem~\citep{yang2023leandojo, song2024largelanguagemodelscopilots} assists users in the task of formalizing a mathematical theorem; Tutor CoPilot~\citep{wang2024tutor} provides real-time guidance to human tutors that teach school students mathematics at the level of math word problems. 

\paragraph{Limits of relying on a single formal system.}
Interactive theorem provers such as Lean provide machine-checkable correctness, and stronger automation inside these systems is valuable. However, there are limits to what is natural to express and maintain inside a single formal foundation and library, even when the underlying mathematics is standard. The following non-exhaustive examples illustrate why a mathematical copilot cannot be reduced to a Lean-only workflow.

First, formal developments can incur large overhead when a concept has many surface variants. Patrick Massot has pointed out this issue using the elementary calculus notion of a limit: a naive approach that treats each variant of “limit at a point” as a distinct definition leads to a combinatorial explosion in the number of definitions and routine plumbing lemmas (his talk gives concrete counts and examples)~\citep{massotLimitsTalk}. In practice, mathematicians avoid learning and manipulating hundreds of separate limit definitions by teaching an abstract generator behind them, or by moving to a more uniform formalism (for example, ultrafilter-based formulations). Both approaches can be implemented formally, but they change what is convenient to teach, state, and reuse.

Second, some common mathematical activities are more naturally carried out in informal language. Concise proof exposition, proof summarization, and “zooming out” to communicate the core idea of a long argument are straightforward in natural language, while formal proofs typically expose low-level structure. Producing readable summaries from formal developments is a separate task that requires its own representations and data.

Third, conjecturing and exploration tend to happen before the final formal statement is stable. Early-stage work often involves proposing definitions, testing variants, and revising hypotheses. Doing all of this directly in Lean can add technical overhead at a point where the main goal is to move quickly through candidate formulations.

Fourth, the choice of foundations affects what counts as “the same” object. For example, work in homotopy type theory was driven in part by a refined analysis of equality and equivalence. Informal mathematics routinely treats objects as equal “at the level that matters” in a given theory, even when they are not definitional equalities in a fixed foundational encoding. Formal developments must commit to a specific notion of equality and transport, and that commitment can make some styles of reasoning less direct.

In the course of proving a theorem, one may need to convert between more abstract mathematics and less abstract mathematics. Flexible conversion might prove to be more difficult in Lean.

These limitations do not argue against formalization. They suggest that Lean is one tool in a broader toolbox, and that mathematical copilots will likely need a federation of tools and representations, including automation over informal mathematics, to support rapid iteration in research workflows.

Mathematics, with its rich set of domain-dependent workflows, tools, and modes of thinking, requires copilots with exceptionally broad capabilities. A mathematical copilot that is useful to a mathematician must significantly exceed the capabilities of the three exemplary copilots mentioned above, which are systems that are designed to help with a narrow technical set of tasks.

\emph{Our paper is split into three main sections}: In the ensuing Section~\ref{sec: past}, we review past datasets that may have limitations of various kinds. We split these into two further categories, relating to natural language and formal language datasets.

\paragraph{Training corpora vs benchmarks.}
In this paper, the term \emph{dataset} covers two use cases: (i) corpora used for pretraining, fine-tuning, or reinforcement learning, and (ii) held-out benchmarks used to report model performance.
Some collections are used in both ways.
Benchmarks are also sometimes converted into training data, for example by creating instruction-style variants of benchmark problems.
As a result, many of the issues discussed below apply to both training and benchmarking, but they affect different parts of the pipeline.
Difficulty calibration and interaction-mode coverage affect the supervision signal during training and the amount of headroom in a benchmark.
Contamination affects both use cases: benchmark items can appear in training corpora, and benchmark scores can be inflated when test problems, or close variants, leak into training data.
Other issues depend more strongly on the use case.
Binary scoring and the evaluation protocol determine what a benchmark score measures.
For formal mathematics, placeholders (\texttt{sorry}, \texttt{Admitted}), duplication across alternative formalizations, and toolchain provenance affect both the quality of training data and the interpretation of evaluation results.
Table~\ref{tab:train_bench_issue_matrix} summarizes these distinctions.

To give an overview of our contributions in that section, current natural language 
datasets typically model a question-response interaction, such as posing a problem and writing a solution. 
We identify dimensions where such datasets frequently fall short of being able to assess advanced mathematics:
\begin{itemize}[leftmargin=*]
    \item \textbf{Difficulty}: Most datasets %
    are situated on a level of difficulty at or below the level of an undergraduate degree in mathematics, or impossibly hard, with few datasets at an \emph{appropriate} level of difficulty, from which model creators can better learn about the strengths and weaknesses of their system.
    \item \textbf{Binary evaluation}: Benchmarks typically only support assessing \emph{if} the output was correct or incorrect, but in case of errors, do not provide information about \emph{how} the output failed to be correct, which is crucial to support.
    \item \textbf{Lack of trusted automation evaluation}: Benchmarks rely on checking a final answer (generally, by keyword matching), imperfect model-based evaluation, or provide no automatic evaluation at all.
    \item \textbf{Standardized Interaction Modes}: Often, natural language datasets only contain question-answer mathematical problems. The daily experience of a mathematician is much richer than that~\citep{zhang2023cogsciMath}, since to prove/search for a theorem, all kinds of \enquote{lateral} modes of thinking have to be applied. Examples are intermediate conjecture formulation, (counter)example search, and finding analogies with other theorems. These are typically not found in textbooks, so there is no representation of these in terms of data.  We point to \citet{collins2023evaluating} for an exposition of the problems of static evaluation for natural-language mathematics problems in practice. In short, static evaluation relies on fixed input–output pairs that fail to capture  iterative or adaptive interactions, such as clarifying questions and corrections, that are an integral part of effective problem solving.
    \item \textbf{Contamination}: Many datasets released solely for evaluation and benchmarking purposes have a short half-life, as, once released, they are easily leaked into datasets that will be used as training data for future models.
    \item \textbf{(Binned) Sankey Diagrams}: Most benchmarks are snapshots of current models, and don't record clearly how performance is changing. This makes it hard to assess the broader direction of evolution of language models w.r.t. mathematics. (Binned) Sankey Diagram offer an easy remedy, but have not been embraced by the community.
    
\end{itemize}

Several of these issues are also present for formal mathematics. But benchmarks around formal mathematics frequently suffer from distinct issues than those listed above; for instance, while datasets for formal mathematics come at a range of difficulties, from competition-level problems~\citep{zheng2021minif2f} to undergraduate textbooks~\citep{azerbayev2023proofnet}, to some results that are at the mathematical state-of-the-art~\citep{scholze2022liquid, paulson2022schemes}, formalized mathematics evaluation can be easily carried out in a safe and automated manner since proofs can be checked by the ITP, rendering the issue of trusted automatic evaluation irrelevant. On the other hand, the following dimensions are problematic for datasets related to formal mathematics:

\begin{itemize}[leftmargin=*]
    \item \textbf{Tool misalignment}: Various automation tools exist that aid in proving small, intermediate lemmas arising in the formalization of more comprehensive results. This creates incentives to change a proof's structure to maximize the use of these tools, potentially resulting in less readable and \enquote{human looking} proofs. This misalignment often manifests as proof-script scaffolding, such as broad tactics applied repeatedly, large automation argument lists, or redundant steps that make proofs brittle and non-idiomatic. Since such redundancy is frequently detectable by linters (e.g.\ unused \texttt{simp} arguments), it can be treated as dataset metadata in corpora that include a raw view for realism along with a linter-cleaned view that better reflects canonical proof structure.
    \item \textbf{Non-trivial data duplication}: In mathematics, it is common that one theoretical concept has several formal representations. For example, the concept of a graph can be described formally in multiple ways. Formal libraries typically feature a variety of representations for the same mathematical concept, some being almost exact copies of each other. Such near-duplication can lead to downstream issues when the library is both used as training and test data in machine learning approaches. Duplication also arises at the level of lemma interfaces: when a library lacks a lemma in the form needed, developers introduce thin wrapper lemmas or local variants to patch an API gap. These near-duplicates are often engineering artifacts rather than conceptual novelty, and they can inflate the overlap between training and evaluation in ways that are invisible to string-based de-duplication.
    \item \textbf{Standardized interaction modes}: (Analogous to natural-language, but with different characteristics) Current formal benchmarks focus on proving theorems in isolation or translating individual statements into formal code. These tasks do not capture all aspects of advanced mathematical practice. In practice, formalization is dominated by interactive patterns that are mostly absent from current benchmarks, including iterative repair (compile $\rightarrow$ inspect diagnostics $\rightarrow$ patch), systematic search for lemmas or imports, and refactoring for readability and performance. For example, Lean supports interactive lemma search via \texttt{exact?}, \texttt{apply?}, and \texttt{rw?}, which reflect a workflow that interleaves searching and proving, rather than proof synthesis from scratch.
    \item \textbf{Scope of formalization and handling of incomplete proofs}:
Formal-math corpora vary substantially in \emph{scope} and \emph{granularity}.  Many benchmarks emphasize short, isolated theorems or textbook-style exercises~\citep{zheng2021minif2f,azerbayev2023proofnet}, whereas large-scale formalization projects capture long-form arguments that unfold across hundreds of intermediate lemmas before reaching a major theorem~\citep{scholze2022liquid,paulson2022schemes}.  This matters for dataset construction and evaluation: models trained or tested only on short lemmas may appear strong while still failing to manage the dependency structure, planning, and bookkeeping required for long developments.

A second, closely related issue is \emph{incompleteness during development}.  In Lean, unfinished results are often recorded using \texttt{sorry}, and in Coq using \texttt{Admitted}; these placeholders accelerate library growth but may render a corpus logically incomplete (and, in the limit, inconsistent).  If such items are included unmarked in training data, a model observes theorems treated as “proven” without ever seeing a proof term, which can distort learning and confound evaluation.  This phenomenon is common in actively developed libraries; one example is SciLean~\citep{SciLean}.  Consequently, formal-math datasets should explicitly document how placeholders are handled (e.g., excluded, retained with flags, or separated into a distinct split), and should report basic completeness statistics (e.g., number or fraction of theorems containing placeholders, and where they occur in dependency graphs).

Finally, to better reflect real mathematical practice, there is value in \textbf{long-form curated corpora} where an entire paper or chapter is represented as a structured development: (1) statements, (2) complete machine-checked proofs when available, (3) dependency metadata, and (optionally) (4) high-level commentary about proof strategy and lemma roles.  Even when full completion is infeasible, clearly distinguishing finished proofs from partial ones enables more meaningful training objectives (e.g., proving the remaining \texttt{sorry} lemmas) and more reliable benchmarks.

\item \textbf{Version fragility and tactic stability}:
Formal proofs are executable artifacts whose replay can be brittle under prover or library drift.  A common failure mode is tactic instability. For instance, in Lean a ``flexible'' tactic such as \texttt{simp} changes the goal in a context-dependent way, and downstream steps silently assume a particular post-\texttt{simp} shape (e.g.\ after simplification one must start a \texttt{calc} chain from the \emph{actual} goal state, not the expected form).

\item \textbf{Hidden context dependence (imports, local instances, and attribute state)}:
Even with a fixed statement, the behavior of a proof script can depend on surrounding environment, including imports, local instances, and attribute state (e.g.\ \texttt{simp} lemmas). In Lean, instance ``pollution'' can occur when local bindings introduce alternate instances (such as multiple \texttt{MeasurableSpace} structures), causing later lines to elaborate against an unintended instance unless the intended instance is explicitly pinned.

\item \textbf{Resource sensitivity (heartbeats and elaboration/typeclass search)}:
Correctness is not cleanly separable from performance: proofs can fail due to heartbeat limits or timeouts caused by elaboration or typeclass synthesis rather than mathematical content. This is especially salient in large developments where small refactors or additional instances can cause elaboration to take much longer, possibly more than the previous heartbeat limit.

\item \textbf{Redundant scaffolding and non-canonical proof artifacts}:
Formal corpora often contain redundant tactic steps and linter-detectable redundancy (such as unused automation arguments, no-op steps) that are harmless for one-off development but distort the signal for training and evaluation. For instance, a tactic may already close the goal, making subsequent steps ill-typed (e.g.\ leading to a Lean error ``no goals to be solved''). A useful dataset practice is to include both a raw corpus and a lint-cleaned canonicalized view.

\item \textbf{Toolchain and automation}:
Many formal corpora are now partially produced by automated or semi-automated methods—ranging from autoformalization tools that attempt to translate informal text into formal proofs \citep{jiang2022draft}, to specialized rewriting tactics, to code-generation scripts that unify definitions across libraries. Current datasets may not have enough detail about which automated methods were used and how their outputs were curated. A related form of provenance is linter metadata: Mathlib-style linters flag redundant arguments, unstable patterns, and stylistic violations, and recording which lints were applied (or waived) can help downstream users understand how canonical the resulting proof scripts are. Similarly, explicit \texttt{set\_option} declarations (e.g.\ heartbeat or recursion-depth overrides) should be treated as dataset metadata, since proofs that require non-default resource limits may behave differently under standard evaluation harnesses.

\item \textbf{Relation of dataset to existing formal libraries}:
Many formal datasets build upon large existing libraries (e.g.\ \texttt{mathlib} in Lean).  If a new dataset reuses or extends such a library, overlaps or version discrepancies can arise.  Two projects may formalize the same theorem in slightly different ways, or rely on different versions of the library's foundational definitions. Version fragility is particularly acute for automation-heavy proofs, as changes to typeclass instance priorities, simp-lemma sets, or default tactic behavior can cause a proof that worked in one commit to fail silently or timeout in the next, even when the underlying mathematics is unchanged.

\item \textbf{Unclear benchmarking protocols}:
Formal mathematics datasets are frequently used as benchmarks for automated theorem proving and related model-based formal reasoning. However, papers often report a single success rate without specifying the task definition and evaluation protocol in enough detail to make results comparable.

At minimum, a formal-math benchmark should state which task is being evaluated. Common variants include:
(i) \emph{theorem proving from scratch} (given a statement and imports, produce a proof that checks),
(ii) \emph{tactic prediction} (given a proof state, predict the next tactic or a short tactic script),
(iii) \emph{proof repair or hole filling} (complete a partially written proof term or script),
(iv) \emph{premise selection} (select a set of lemmas from a library that suffices to close a goal),
and (v) \emph{autoformalization} (translate an informal statement into a formal one, optionally followed by proving it).

Even within a single task label, the protocol can change difficulty by orders of magnitude. For reproducibility, benchmark authors should specify:
\begin{itemize}[leftmargin=*]
    \item \textbf{Input specification}: what the model receives (the statement only, or also local context such as imported modules, available lemmas, and current proof state), what implicit environment is assumed (active typeclass instances, scoped attributes, and simp-lemma sets), and any context limits (number of retrieved premises, maximum file size, or maximum tokens).
    \item \textbf{Output specification}: what counts as a valid submission (a full proof script, a proof term, a sequence of tactics, or a patch to an existing file).
    \item \textbf{Success criterion}: whether the artifact must be accepted by the checker without placeholders (e.g., no \texttt{sorry}/\texttt{admit}), and what counts as failure (type error, timeout, unsolved goals).
    \item \textbf{Tooling and automation}: which tactics and automation are permitted (e.g., rewriting and simplification tactics, SMT calls, hammer-style tools), and whether tool use is counted as part of the model or treated as an external oracle. If multiple settings are relevant, separate tracks (for example, ``with automation'' and ``restricted automation'') avoid conflating results.
    \item \textbf{Resource constraints}: wall-clock time per goal, memory limits, maximum number of proof attempts, and any search limits (tactic budget, backtracking depth). Limits that are internal to the prover matter as well: heartbeat budgets, typeclass-synthesis depth, and recursion limits can all cause a correct proof script to fail silently or timeout when replayed under different settings. These choices interact strongly with inference-time compute and must be reported.
    \item \textbf{Environment pinning}: the exact proof assistant version and library commit (for example, Lean + mathlib revision, or Isabelle + AFP snapshot), since minor version changes can make proofs succeed or fail. An evaluation harness that replays proofs in a fixed environment reduces ambiguity.
    \item \textbf{Metrics}: beyond a single success rate, report pass@$k$ when multiple attempts are allowed, average time to solution, and optional secondary measures such as proof length or number of tactic steps, since these affect usability.
\end{itemize}

Without these details, two papers can report the same score while solving different tasks under different constraints. Explicit protocols and pinned environments enable reliable comparison across models and make it possible to interpret gains as improvements in formal reasoning rather than changes in tooling or evaluation setup.

\end{itemize}

\subsubsection{Mixed-mode datasets for tool use}
\label{subsec:mixed_mode}

Many candidate mathematical copilots are not text-in, text-out systems. They are trained and evaluated in settings where a model is expected to call external tools, such as a computer algebra system, a numerical solver, a proof assistant, a retrieval system, or a specialized symbolic model. Tool-integrated systems already appear in work on tool-integrated reasoning for math (for example, models that generate and execute Python code as part of the solution process~\citep{gou2023tora}).

A \emph{mixed-mode dataset} records these interactions explicitly. Rather than storing only a problem statement and a final response, each datapoint is a short trajectory consisting of interleaved representations, for example:
(i) natural-language problem context,
(ii) intermediate formal objects (expressions, goals, proof states, or formal statements),
(iii) tool calls with arguments,
(iv) tool outputs (for example, the result of executing code, a CAS simplification, or a proof assistant error message),
and (v) a final response that integrates the tool outputs into a mathematically coherent explanation.
This structure supports training objectives beyond next-token prediction, including learning when to call a tool, how to translate between representations, and how to recover from tool failures (timeouts, type errors, missing imports, or non-terminating proof searches).

A special case of mixed-mode data consists of datasets pairing \emph{informal} and \emph{formal} mathematics. These datasets support autoformalization and naturalization, as well as theorem proving conditioned on an informal statement. Examples include miniF2F~\citep{zheng2021minif2f} and ProofNet~\citep{azerbayev2023proofnet}. More recent efforts scale paired informal-formal corpora to substantially larger sizes~\citep{ying2024lean}, which matters for training rather than only evaluation. In these settings, the tool is the proof assistant itself: correctness can be checked by verifying the generated formal artifact, while intermediate feedback takes the form of proof-state information and type errors, which can be recorded in the dataset.

Mixed-mode datasets introduce evaluation and reproducibility issues that do not arise in static text-only benchmarks. Tool outputs depend on the toolchain (version, library state, and sometimes randomness). A dataset intended for benchmarking should therefore specify the tool interface and pin versions (for example, by commit hashes for proof assistant libraries), and provide an evaluation harness that can replay tool calls in a controlled environment. Without this, differences in tool versions can change whether a proof checks or whether a symbolic computation succeeds, making scores hard to compare across papers.

\paragraph{Recommendations.}
For mixed-mode datasets used for training or benchmarking, we recommend (1) representing each datapoint as an explicit tool-interaction trace, including tool inputs and outputs; (2) recording the toolchain in sufficient detail to replay the trace (versions, dependencies, and execution settings); and (3) separating tasks where tool use is optional from tasks where tool use is required, so that evaluation can distinguish mathematical reasoning from tool invocation competence. Mixed-mode datasets also interact with the contamination problem: if tool outputs encode the final answer too directly, a model can learn to pattern-match on tool outputs rather than learning to select and use tools effectively. This motivates evaluation protocols that check both the final answer and the correctness of the tool trace itself (for example, by re-executing code or re-checking formal proofs).

\begin{table}[t]
\centering
\small
\renewcommand{\arraystretch}{1.15}
\resizebox{\linewidth}{!}{
\begin{tabular}{p{6.2cm}cccc}
\arrayrulecolor{gray!15}
\rowcolor{gray!40}
\textbf{Issue} & \textbf{NL Train} & \textbf{NL Bench} & \textbf{Formal Train} & \textbf{Formal Bench} \\
\rowcolor{gray!15}
Difficulty calibration and coverage & X & X & X & X \\
Binary scoring / success-only reporting & X & X & X & X \\
Lack of trusted automatic evaluation & X & X &  &  \\
Standardized interaction modes (Q-A, isolated lemmas) & X & X & X & X \\
Contamination and leakage & X & X & X & X \\
Binned progress reporting (e.g., Sankey-style) &  & X &  & X \\
Tool misalignment from automation / tactic availability &  &  & X & X \\
Non-trivial duplication (multiple representations / near-copies) &  &  & X & X \\
Incomplete proofs and placeholders (\texttt{sorry}, \texttt{Admitted}) &  &  & X & X \\
Toolchain and automation provenance (scripts, autoformalization, tactics) &  &  & X & X \\
Dependence on external libraries and versioning (commits, environments) &  &  & X & X \\
Unclear benchmarking protocol (timeouts, allowed tools, environment pinning) &  & X &  & X \\
\end{tabular}}
\caption{Issue matrix for datasets used as training corpora (Train) and as benchmarks (Bench), separated by natural language (NL) and formal mathematics. ``X'' indicates the issue commonly or sometimes affects that use case. For formal mathematics, proof checking provides automated correctness, so the row ``Lack of trusted automatic evaluation'' is typically not applicable.}
\label{tab:train_bench_issue_matrix}
\end{table}

These limitations have various consequences: For example, evaluating progress of the mathematical capabilities of AI systems, particularly for advanced mathematics, is difficult. 
For instance, improved accuracy on a standard problem-solving dataset may not provide a meaningful signal about what has improved, while performing well on an existing formal benchmark may not accurately indicate where models currently struggle. 

In the next section, Section~\ref{sec: future}, we discuss how certain aspects of the proof creation can be mapped to datasets. We argue that such datasets that describe intermediate stages that mathematicians go through in the long process of devising proofs are potentially highly valuable to support mathematical copilots that assist humans in these tasks. The overt focus of current datasets on \emph{result} (the final proof), rather than the process that led to that proof, severely diminishes their usefulness for training LLMs to be used as mathematical copilots. While from a perspective of pure proof search, results-based datasets are justified, it is unlikely that mathematics will soon be reduced purely to proof search, as there are several mathematical activities (such as finding the \enquote{right} set of definitions, that makes a mathematical theory clear and easy to follow) that are not solely about proof search.
Such datasets that describe these intermediate stages do not exist at all currently! The GHOSTS dataset~\citep{frieder2023mathematical}, to our knowledge, is the only one that preliminarily investigates how well LLMs can engage in literature search. Thus, in Section~\ref{subsec: workflows}, we take a more holistic level to describe how mathematical workflow might be described by datasets. This problem is essential, as a lot of mathematical activity is based on chaining together complex workflows. There is currently no clear way of measuring the advances of AI systems in tackling the full spectrum of mathematical workflows. Such data about intermediate stages in the proof discovery process is often not explicitly contained in textbooks and articles -- the data sources on which current LLMs are trained. In Section~\ref{subsec: data collection}, novel ways of capturing data that may contain this information are described. Finally, in Section~\ref{sec: motivated proof}, we discuss a stricter proof structure called a \enquote{motivated proof}, requiring the author to explain the origin behind each step of a proof. We then argue for an evaluation benchmark enforcing this proof structure from the perspectives of end-user utility and evidence of LLM reasoning rather than LLM memorization (coming up with a motivated proof, given an unmotivated one, requires reasoning and deeper proof understanding) and speculate on paths toward designing such a benchmark.

We note that in this article we do not focus on datasets that are designed primarily to be consumed by the current LLM technology stack, for example, various instruction tuning datasets, such as the OpenMathInstruct dataset~\citep{toshniwal2024openmathinstruct}. These are often based on existing datasets (GSM8K and MATH in the case of OpenMathInstruct) and thus could be described as \emph{derived} datasets. Rather, our focus is on assessing how much and how well \emph{primary} datasets describe various aspects of \emph{doing mathematics}.

\begin{table}
\centering
\renewcommand{\arraystretch}{1.2}
\resizebox{\linewidth}{!}{\begin{tabular}
{llrp{4.5cm}p{8.25cm}}
\arrayrulecolor{gray!15}
\rowcolor{gray!40}
\textbf{Dataset} & \textbf{Reference} & \textbf{Size} & \textbf{Field} & \textbf{Description} \\
\rowcolor{gray!15}
\multicolumn{5}{c}{\textbf{Elementary to Middle School Level}} \\
\textbf{Commoncore / MultiArith} & \citet{roy2015solving} & 600 & Basic arithmetic & Word problems with two steps \\
\hline
\textbf{MAWPS} & \citet{koncel2016mawps} & 3320 & Basic arithmetic and algebra & Word problems requiring systems of up to two linear equations (includes MultiArith) \\
\hline
\textbf{Math23K} & \citet{wang2017deep} & $23161$ & Basic arithmetic & Word problems in Chinese \\
\hline
\textbf{ASDiv} & \citet{miao2020diverse} & $2305$ & Basic arithmetic & Word problems (selected from the internet) \\
\hline
\textbf{SVAMP} & \citet{patel2021nlp} & $1000$  & Basic arithmetic & Word problems requiring up to two steps (derived from ASDiv)  \\
\hline
\textbf{GSM8K} & \citet{cobbe2021training} & $8500$ & Basic arithmetic & Word problems requiring 2-8 steps to solve \\
\hline
\textbf{GSM8K-symbolic} & \citet{mirzadeh2024gsm} & $5000$ & Basic arithmetic & Variants of GSM8K problems using symbolic templates \\
\rowcolor{gray!15}
\multicolumn{5}{c}{\textbf{High School to Intermediate Level}} \\
\textbf{Dolphin18K} & \citet{huang2016well} & $18460$  & Basic arithmetic and algebra & Problems and answers from the mathematics category of \emph{Yahoo! Answers} \\
\hline
\textbf{AQuA} & \citet{aquarat2017ling} & $100000$ & Basic arithmetic and algebra & Algebraic multiple-choice word problems (similar to GMAT and GRE)  \\
\hline
\textbf{MATH} & \citet{hendrycks2021measuring} & $12500$ & Problem-solving & Problems from high school math competitions with step-by-step solutions \\
\hline
\textbf{NumGLUE} & \citet{mishra2022numglue} & $101835$ & Arithmetic reasoning & 8 different mathematical reasoning tasks (including previous datasets). \\
\hline
\textbf{GAOKAO} (Math) & \citet{zhang2023evaluating} & $844$ & Various high school math topics & Questions from the \emph{Chinese College Entrance Exam} (single-choice, cloze, open-ended) \\
\hline
\textbf{Hungarian High School Finals} & \citet{testing_language_models_on_a_held_out_high_school_national_finals_exam} & $33$ & Problem-solving & Problems from the 2023 \emph{Hungarian National High School Finals} in mathematics \\
\hline
\textbf{MATH 401} & \citet{yuan2023well} & $401$ & Advanced arithmetic and computation & Arithmetic expressions and numerical results \\
\hline
\textbf{TAL-SCQ5K} & \citet{tal-scq5k} & 5000 & Various & Multiple-choice competition questions from elementary to high school levels in English and Chinese  \\
\hline
\textbf{AGIEval} (Math) & \citet{zhong2024agieval} & $1943$ & Various & Problems (in English and Chinese) derived from college admission tests (GRE, Gaokao, SAT) and high school competitions (from MATH and AQuA) \\
\rowcolor{gray!15}\multicolumn{5}{c}{\textbf{College Level}} \\
\textbf{TheoremQA} & \citet{chen2023theoremqa} & $800$ &  Various (including algebra, number theory, graph
theory, information theory) & Questions (from the internet and textbooks) and expert answers based on $354$ theorems \\
\hline
\textbf{ARB} & \citet{sawada2023arb} & $234$ & Various & Problems from university exams and competitions at the undergraduate level  \\
\rowcolor{gray!15}
\multicolumn{5}{c}{\textbf{Graduate to Research Level}} \\
\textbf{NaturalProofs} & \citet{welleck2021naturalproofs} & $25271$  & Various (focus on algebraic geometry) & Theorems with proofs and reference graphs from \emph{ProofWiki}, the \emph{Stacks project}, and textbooks. \\
\hline
\textbf{GHOSTS} & \cite{frieder2023mathematical} & $709$ & Various & Problem-solving, theorem-proving, and retrieval tasks derived from textbooks, \emph{StackExchange}, and existing datasets (MATH and Symbolic integration) \\
\hline
\textbf{FrontierMath} & \citet{glazer2024frontiermath} & \hspace{-0.5em}$>35$ (unknown) & Various & Challenging problems with automated verification created by mathematicians \\
\rowcolor{gray!15}
\multicolumn{5}{c}{\textbf{Olympiad Level}} \\
\textbf{miniF2F} & \citet{zheng2021minif2f} & $488$ & Problem-solving, algebra, number theory & Formalized Olympiad-type problems from AIME, AMC, IMO, as well as high school and undergraduate courses (also derived from MATH) \\
\hline
\textbf{OlympiadBench} & \citet{he2024olympiadbench} & $8476$ & Various & Problems and solutions from mathematics and physics competitions and the \emph{Chinese College Entrance Exam} (in English or Chinese and partially with images) \\
\hline
\textbf{IMO Small Challenge} & \citet{frieder2024imo} & $100$ &  Combinatorics & Selected problems and solutions from IMO and BWMC \\
\hline
\textbf{IMO-AG-30} & \citet{trinh2024solving} & $30$ & Geometry & Plane Euclidean geometry problems from IMO in JGEX formal language \\
\rowcolor{gray!15}
\multicolumn{5}{c}{\textbf{Various / Mixed Difficulty}} \\
\textbf{Mathematics} & \citet{saxton2019analysing} & Generated  & Arithmetic, algebra, probability, calculus & Sequential questions generated with modular structure \\
\hline
\textbf{MMLU} (Math) & \citet{hendrycks2020measuring} & $960$  & Elementary, high school, and college mathematics, and abstract algebra & Multiple-choice questions collected from online sources ranging from elementary to college (similar to GRE) mathematics \\
\hline
\textbf{INT} & \citet{wu2020int} & Generated  & Inequalities, General Mathematics & Generated formalized (in-)equality theorems based on ordered field
axioms \\
\hline
\textbf{AMPS} & \citet{hendrycks2021measuring} & $\approx 5100000$ & (Linear) algebra, calculus, statistics, geometry, 
and number theory & Problems and step-by-step solutions from Khan Academy ($>100$k) and generated using Mathematica ($\approx 5$M) \\
\hline
\textbf{Lila} & \citet{mishra-etal-2022-lila} &  132239  &  Arithmetic, (linear) algebra, calculus, statistics, geometry, 
and number theory & Derived from $20$ existing datasets (including NumGLUE, SVAMP, MultiArith, Dolphin18K, Mathematics, AMPS, GSM8K, MATH) with numbers or expressions as outputs \\
\hline
\textbf{CMMLU} (Math) & \citet{li2023cmmlu} & $499$ & Various & Multiple-choice questions (in Chinese) from elementary to college level collected from freely available resources \\
\hline
\end{tabular}}
\caption{Selected datasets for evaluating the mathematical capabilities of LLMs.}
\label{tab:math_datasets}
\end{table}

\section{The Past -- Common Pitfalls in Current Datasets}
\label{sec: past}

In this section, we discuss several issues with current datasets. Note that our limitations are with respect to the landscape of AI for mathematics at the time of writing (Dec 2024); it is possible that some of these limitations are addressed by the time the reader reads this article. Indeed, we hope that they \textit{are} addressed swiftly. 

Underlying several of these problems is a dynamic described by Goodhart's law: ``When a measure becomes a target, it ceases to be a good measure.'' A related formulation, Campbell's law~\citep{campbell1979assessing}, warns that quantitative indicators used for decision-making become increasingly subject to corruption pressures. In the context of mathematical AI, when LLM developers optimize for specific benchmark scores (such as accuracy on MATH or GSM8K), models may learn to exploit superficial patterns or benefit from data contamination rather than developing robust mathematical reasoning. High benchmark scores then no longer reliably indicate the capabilities that matter for mathematical copilots. This motivates our call for richer evaluation frameworks that are harder to ``game'' and that measure diverse facets of mathematical competence. Notably, the NLP community identified these same issues years ago and proposed solutions such as dynamic, adversarial benchmark creation~\citep{kiela-etal-2021-dynabench}, yet the math-AI community has largely continued to rely on static benchmarks susceptible to the same failure modes.

Recent near-saturation results on boxed-answer benchmarks~\citep{deepseek2025r1,qwen2024qwq32bpreview} are consistent with this dynamic: as optimization concentrates on a small set of public targets, benchmark scores become less informative about broader mathematical competence.

We first list issues that are common to both natural-language datasets and datasets comprising formalized mathematics. Then, we list issues that are specific to either natural-language datasets or formal datasets.

\subsection{Common Issues}

\subsubsection{Difficulty}
\label{subsubsec:difficulty}
The difficulty of mathematical problems typically varies along at least two salient dimensions: level of abstraction and proof (or solution) sophistication. Abstraction varies from simple \enquote{one-layer} definitions, such as those that involve properties of topological spaces, to those in areas such as algebraic geometry, where definitions of highly intricate mathematical objects involve several layers.

Proof sophistication is related to problem-solving ability and is often tested (for humans and AI systems alike) in mathematical competitions, such as the International Mathematical Olympiad (IMO). Typical for these is that the statements to be proved are elementary, but highly ingenious proof techniques need to be used to arrive at the solution.
 
Several datasets have been proposed which explore proof sophistication, such as the OlympiadBench~\citep{he2024olympiadbench}, the IMO Small Challenge, or the IMO-AG-30 dataset~\citep{trinh2024solving}, but very few existing datasets reach a level of difficulty that is more advanced than that of an undergraduate degree in mathematics in terms of level of abstraction. For a (non-exhaustive) overview of datasets, ordered by their difficulty, see Table~\ref{tab:math_datasets}.

For example, MATH focuses on high-school competition problems, while the recent TheoremQA~\citep{chen2023theoremqa} and Advanced Reasoning Benchmark (ARB)~\citep{sawada2023arb} are upper-undergraduate level.
Many commonly used datasets test mathematics below the undergraduate level, such as GSM8k and the 23 tasks in the Lila benchmark~\citep{mishra-etal-2022-lila}. 

Two datasets that go beyond upper-undergraduate level are the GHOSTS dataset~\citep{frieder2023llms}, as well as the NaturalProofs(-Gen)~\citep{welleck2021naturalproofs,welleck2022naturalprover} datasets, which contain problems, theorems, and proofs on a wide range of more advanced mathematical topics.
However, these suffer from a lack of automated evaluation and contain many problems that could, in principle, be in the training corpora of modern LLMs, as we discuss later. FrontierMath~\citep{glazer2024frontiermath} is also a more advanced dataset, but is currently not publicly accessible. STORM-BORN is a recent derivation-style benchmark curated from recent research papers; it reports that strong models solve only a small fraction of the selected problems without task-specific training~\citep{stormborn2025}.

Most current LLMS (e.g., the Qwen~\citep{yang2024qwen2} family of models, or commercial models released by OpenAI and Google) are evaluated on the MATH dataset~\citep{hendrycks2021measuring} and the GSM8K dataset~\citep{cobbe2021training}, with a recent trend of some LLMs using held-out datasets such as the Hungarian National Finals high-school exam as an additional test of generalization~\citep{testing_language_models_on_a_held_out_high_school_national_finals_exam}. 
The MATH dataset uses five categories of difficulty and is strictly more difficult than GSM8K. Nonetheless, state-of-the-art LLMs have been found to reach an acceptable performance on it~\citep{frieder2023mathematical}. 
For instance, Gemini reaches a performance of about 50\%.\footnote{\url{https://blog.google/technology/ai/google-gemini-ai/performance}}, while solving almost all problems on GSM8K. Progress in LLMs is proceeding at a rapid pace, and the \enquote{Math-Specialized} version of Gemini 1.5 reaches (at best) a score of 91\% on MATH, which is representative of the current state of the art (although, due to the closed nature of this line of research, it is difficult to place this achievement into context, such as by analyzing how similar training data has been to the MATH dataset, and how much was needed). 

While this performance may appear impressive, neither of these datasets covers a high degree of difficulty in terms of abstractness and proof sophistication. Hence, the current performance of LLMs on these datasets is not yet indicative of advanced mathematical understanding, which is relevant for professional mathematicians.

This is particularly striking in the case of AlphaGeometry, which can solve 25 geometry problems from the IMO out of a test set of 30 problems~\citep{trinh2024solving} (and 213 out of 231 on a larger benchmark). We argue that the accompanying dataset, while indeed difficult in terms of proof sophistication, is not sufficiently large to be able to accurately compare AlphaGeometry with other models.

Moreover, recent work casts doubt on whether high benchmark scores reflect genuine mathematical reasoning. \citet{nezhurina2024alice} demonstrate that GPT-4, Claude 3 Opus, and other state-of-the-art LLMs at the time, exhibit dramatic reasoning failures on simple common-sense math problems (the ``Alice in Wonderland'' or AIW problems) that are trivially solvable by humans. These failures persist even with chain-of-thought prompting and multi-step re-evaluation, and models express strong overconfidence in their incorrect answers. If LLMs struggle with such elementary problems, it is difficult to credit their high performance on substantially harder benchmarks like MATH to genuine reasoning rather than to memorization or pattern-matching on familiar problem types. Proof or Bluff? reports expert grading of model solutions on the 2025 USAMO and finds low proof scores even for models that score well on short-answer benchmarks~\citep{prooforbluff2025}. This tension motivates two directions: first, the development of out-of-distribution evaluation sets --- for instance, via procedural generation of novel problems --- that are less susceptible to memorization; and second, decontamination methods tailored specifically to mathematics, which must account for the fact that mathematically equivalent problems can appear syntactically different.

\paragraph{Recommendations.} We believe that it is crucial to introduce more difficult datasets into the evaluation repertoire of math-AI. And importantly, the skill level of these datasets should be \textit{graded}, such that we can form a nuanced characterization of model capabilities -- this can be achieved, in part, by interpolating difficulty level between existing datasets and those representative of advanced mathematics. This includes datasets on graduate-level mathematical domains, such as functional analysis (to measure handling of arguments involving high levels of abstraction), partial differential equations (to measure handling of intricate computational arguments), as well as other domains that come with their own \enquote{mathematical flavor}. At the same time, more comprehensive datasets illustrating and measuring advanced problem-solving skills are needed. The miniF2F~\citep{zheng2021minif2f} dataset and the GHOSTS dataset are just the first steps in this direction. For natural language datasets, we note that simply scraping many competition-level problem sets, coupled with an evaluation principle that only verifies a \enquote{final answer}, is not sufficient to ensure that systems can progress to becoming useable mathematical copilots. Such problem sets, even if some of them ultimately are only math word problems, require expert, specialized knowledge and training for humans to be proficient. Without a detailed evaluation protocol (see Section~\ref{subsec: binary benchmarks}), not much insight will be gained into why poor performance was obtained - and, conversely, human evaluation will be very costly (see Section~\ref{subsec: auto eval}).

\subsubsection{Binary Benchmarks}
\label{subsec: binary benchmarks}

Most datasets tailored for machine learning, except GHOSTS and NaturalProofs-Gen, including the dataset for AlphaGeometry, use a simple correct-incorrect rating scheme. The given feedback indicates \emph{if} a system fails to solve a problem correctly, but without any indication of \emph{how} it fails.
Binary evaluation thus provides limited information on how to improve or interpret performance.

\paragraph{Benchmark diagnosticity and ``training on the test task.''}
A further limitation of many mathematics benchmarks is that they offer little insight into why a model succeeded or failed, beyond a single correctness bit. In practice, this can make it hard to improve a system other than by training on large quantities of data that is closely matched to the benchmark task distribution. Dominguez-Olmedo et al.~\citep{dominguez2024trainingonthetesttask} describe this as \emph{training on the test task}: unlike leakage or contamination, it is a legitimate use of knowledge about benchmark tasks at training time, but it can confound both relative model comparisons and claims about emergent abilities. In mathematics, the problem is amplified by the concentration of benchmarks on a small set of task formats; Frieder et al.~\citep{frieder2023mathematical} describe a ``vast unpopulated landscape of benchmarks'' beyond the few overstudied datasets. One response is to make benchmarks more diagnostic by attaching structured feedback to failures, such as error codes (as in GHOSTS) that distinguish common mathematical failure modes, which can guide targeted data curation and model improvements rather than generic task-matching fine-tuning.

GHOSTS and NaturalProofs-Gen address this problem by using labels that are attached to language model outputs, which provide a classification of (mathematical) errors. For instance, GHOSTS distinguishes two types of labels, called \enquote{error codes} and \enquote{warning codes}, which make up, in total, 15 error or warning labels, while NaturalProofs-Gen distinguishes between 11 types of errors.
A small number of datasets use non-binary error rating schemes, to our knowledge these are only TheoremQA (3-point scheme) and ARB (5-point). 

Finally, formalized mathematics offers feedback from the interactive theorem prover (ITP) when a system fails to produce a correct proof (one case being timeouts). 
However, this feedback is more akin to a compiler error than high-level feedback on the mathematical reasoning process. We speculate that this feedback may be useful but still limited in its ability to diagnose AI systems.

\paragraph{Recommendations.} We believe that it is crucial to develop datasets that support multi-aspect feedback.
Relying on a single numerical representation, in the face of the diversity of mathematical failure modes, risks providing a signal that is too weak to be suitable for learning or evaluation -- which is especially important to guard against when we consider deploying systems alongside people, who may care not only that a model is correct but also that the response is appropriately helpful~\citep{collins2023evaluating}.

One of the reasons that AI researchers may have avoided multi-aspect feedback, particularly subjective feedback, is for fear of the difficulty of obtaining consistent human annotations. While annotator disagreement is a challenge, we suggest that the math-AI community look to the ways in which computer vision researchers have worked on developing new theoretical and empirical tools which \textit{lean in} to the diversity of human responses~\citep{uma2021learning, collins2022eliciting, sucholutsky2023informativeness, gordon2022jury}. Appropriately interleaving formality and subjectivity -- especially for advanced mathematics, where there is a limited pool of possible human annotators -- is a challenge, but one which demands cross-disciplinary collaboration and is ripe for future work. 
As a middle-ground, we also recommend considering evaluation methods that do not assign numerical ratings at all -- but rather annotate the outputs of AI systems with error codes and warning codes, then derive ratings from the codes directly (where no error codes mean a perfect rating).

\subsubsection{Standardized Interaction Modes}
\label{sec:standardized}

Almost all datasets to date that are designed to evaluate LLMs capture only textbook-like questions. Moreover, owing to issues of automatic evaluation described in~\Cref{subsec: auto eval}, the majority of datasets focus on a smaller subset of questions with a unique, typically numerical, answer. While such questions might span a range of different topics, from numerical (e.g., GSM8K, problems of level 1-2 from MATH, MultArith) and symbolic evaluations (the dataset associated with the work of~\citep{lample2019deep}) to somewhat complex problem-solving tasks (problems of level 4-5 from MATH), they do not come close to covering all aspects of the mathematical tasks a mathematician might encounter in their daily practice. We believe that any future \emph{mathematical assistant} should provide support for these tasks.

In the following list, we describe different dimensions of interaction modes for natural-language mathematics (both question types as well as session types), across which we believe that mathematical assistants should be evaluated.

\paragraph{Question Types.} 
\begin{enumerate}
    \item \textbf{School/University Curriculum-Like Questions} that are encountered in educational settings and have well-defined answers. This includes questions with arithmetic, symbolic content, math word problems, as well as problem-solving and proofs.
    \item \textbf{Proof-Specific Questions} as the concept of a mathematical proof is varied
    and gives rise to a host of specific questions about proofs, which go beyond the type of questions related to proofs from the previous point. These are advanced questions concerning matters such as: Establishing whether a proof is effective or not, in the sense of proving a general statement about a mathematical object merely by proving its existence as opposed to producing a witness or an algorithm that computes; investigating what distinct proof strategies might be used to prove a certain statement; exploring the distinction between whether an \enquote{elementary} or an \enquote{advanced} proof is given. 
For example, the prime number theorem has classical complex-analytic proofs (via properties of the Riemann zeta function)~\citep{hadamard1896distribution} as well as later \enquote{elementary} proofs~\citep{selberg1949elementary,erdos1949newmethod}.
    \item \textbf{Pictorial representation.}
Many mathematical objects and arguments have standard diagrammatic encodings that support reasoning and communication. Examples include plotting functions to inspect qualitative behavior (monotonicity, convexity, asymptotics), diagram-based proofs and explanations (for example, ``proofs without words'' collections~\citep{nelsen1993proofs}), and commutative or string diagrams used in areas such as algebra, topology, and category theory~\citep{selinger2011survey}. A mathematical copilot that supports research practice should therefore be evaluated on tasks that translate between symbolic or textual descriptions and graphical representations, and on tasks that interpret diagrams to extract relevant properties or propose next steps in an argument.

    \item \textbf{Mathematical explanations and intuition}, e.g., by providing (counter-)examples. Such questions are open-ended and have potentially (infinitely) many correct answers. This freeform exploration is crucial in the process of establishing the truth value of a given (advanced) mathematical statement, where one swings between attempting a direct proof of the statement and finding a counterexample to the statement.
    Terence Tao has recounted encountering a counterintuitive statement in compressed sensing, first attempting to construct a counterexample, and then redirecting that effort into a proof \citep{bernstein2019mind}.

     One recent counterexample-focused benchmark is CounterMATH, which asks models to refute undergraduate-level statements by producing explicit counterexamples (1,216 statement-rationale pairs)~\citep{countermath2025}.
\item \textbf{Proficiency in using mathematical tools}, where the assistant is given access to external tools such as computer algebra systems (SageMath, Mathematica, Maple), numerical solvers, and domain-specific packages (for example, Singular, GAP, SnapPy), and is evaluated on four steps: selecting an appropriate tool, translating the mathematical intent into the tool's input language, validating and interpreting the tool's output, and composing multiple tool calls into a coherent mathematical argument. Tool-integrated reasoning approaches already use this pattern in restricted form (for example, delegating substeps to executable code)~\citep{gou2023tora}. This also clarifies why Lean alone does not cover the full workflow of research mathematics: Lean can verify formal proofs, but many routine sub-tasks (symbolic simplification, numeric exploration, combinatorial search, heuristic experimentation) are more naturally handled by specialized systems. A practical copilot therefore needs a federation of tools, each with its own interface and notion of correctness, and datasets should record tool traces so that evaluation can replay the same sequence of calls.

    \item \textbf{Retrieval Tasks} for definitions and mathematical facts, either from name to statement or reverse (\enquote{mathematical search engine}). MIRB provides a benchmark suite for mathematical retrieval, including premise retrieval and formula retrieval, which can serve as components of literature-search workflows~\citep{mirb2025}. Retrieval tasks can range from a well-defined answer (e.g., \enquote{\emph{Define this mathematical object!}}) to open-ended questions (\enquote{\emph{what are the most important theorems in a certain field?; which results are relevant to prove a certain statement?; has a form of this statement already been proved?}}). This tests the capabilities of mathematical assistants to provide high-level explanations similar to those of a human expert in a certain mathematical field.
    \item \textbf{Informal Proof Completion}, where a proof (or generally a mathematical statement) needs to be checked or completed. In the best case, the mathematical assistant should be able to provide a correction and fill in gaps. In such functionality, mathematical assistants can be running in the background to act as \enquote{copilots} and \enquote{mathematical grammar-checkers} --- as the current generation of copilots\footnote{Such as GitHub's Copilot: \url{https://github.com/features/copilot}} does successfully for programmers.
    \item \textbf{Mixed-mode interactions}, where translation between formal and informal mathematics needs to be performed. These are particularly important for the combination of ITPs and general-purpose LLMs, which has recently picked up interest (e.g.,~\citep{jiang2022draft, first2023baldur,frieder2023llms}). 
\end{enumerate}

To our knowledge, GHOSTS is the only natural-language dataset that focuses on all of the first four items from above. Accompanying this variety of questions (\emph{what} is asked) are session types that describe \emph{how} the questions are posed. 

\paragraph{Session Types.}
\begin{enumerate}
    \item \textbf{Zero-Shot}, where the model is prompted without further information.  
    \item \textbf{In-Context Learning}, where additional background information is provided, ranging from clarifications of the notation to context (e.g., the chapter of the book or paper), to similar questions with corresponding proof. 
    \item \textbf{Interactive}, where the prompter (whose mathematical abilities can span a wide spectrum) solves mathematical questions together with the language model in an interactive way as in~\cite{collins2023evaluating}. 
\end{enumerate}

In practice, typical mathematical workflows require a mixture of different question and session types. 
Even if a question is not open-ended, the amount of mathematical detail and rigor the answer should contain will depend on the user's background. In part, this should be deduced from the way the question is phrased. Currently, no datasets offer support for this.
We note that with the exception of \cite{collins2023evaluating} and the collection of CheckMate, almost all datasets are zero-shot. Interactive training modes are extensively employed in a formal language setup. %
IMProofBench evaluates research-level problems in an agentic setting with tools (web search and computer algebra), combining automated scoring on subproblems with expert grading of full proofs~\citep{improofbench2025}.

For formal mathematics, different considerations apply, as outlined below. While modern LLMs can, to a degree, perform the tasks below, they are not specialized for these tasks:
\begin{itemize}
    \item \textbf{Refactoring} formal mathematics to yield nicer-looking proof is a significant undertaking. A tool that takes as input a formal proof and refactors it along desired criteria would be a welcome addition to the toolbox of researchers working in formalizing mathematics. To our knowledge, the only work to date that attempts this is the \emph{ImProver}~\citep{ahuja2024improver} for Lean, which allows formal proof creation of specific length.
    \item \textbf{Interoperability} between different libraries of formal proofs (such as Lean's Mathlib library and Isabelle's AFP) is a long-term (but still largely unachieved\footnote{A score of overlap by assessing how many theorems out of 100 theorems have been formalized in different ITPs is provided here: \url{https://www.cs.ru.nl/~freek/100/}}) goal. Currently, it takes intimate knowledge of the various formal libraries in order to assess which library, and ensuing ITP, would best be used to formalize a given mathematical theorem. A tool that can assess the suitability of a certain ITP and its library before starting the formalization process could make formalization more accessible.
\end{itemize}

\paragraph{Recommendations.} We advise exploring more diverse mathematical interaction modes -- across a variety of question and session types -- as outlined here, both for formal as well as informal datasets. Our list is not exhaustive and can also be broken down to a more detailed level. In particular, oral, mathematical communication is not well mapped to data (see Section~\ref{subsec: data collection}). Mathematical collaboration often involves an exchange of very high-level ideas at a rapid pace. Specific, wide-ranging recommendations can be found in Section~\ref{sec: future}. We believe capturing this in data will help to lead to more widespread mathematical assistants. 
Finally, we believe that input from research in mathematical education must be taken into account to derive curated datasets that exhibit richer interaction modes.

\subsubsection{Contamination}

When releasing a dataset, control is lost over whether the dataset ends up as training data for machine learning models. For several of the state-of-the-art models, such as GPT-4~\citep{openai2023gpt4}, that perform well on benchmarks, no information is available in the ensuing on whether efforts were made to ensure that training datasets were decontaminated and no datapoints from the evaluation benchmark were included in the training set. For Gemini 1.5~\citep{reid2024gemini}, Qwen2~\citep{yang2024qwen2} and DeepSeekMath~\citep{shao2024deepseekmath}, only $n$-gram decontamination approaches have been tested. For complex datapoints, such as math, $n$-grams are unfortunately not always sufficient to ensure the training dataset is clean. There is evidence that suggests that several such modern models are already contaminated~\citep{xu2024benchmarking}. While mitigating approaches have been proposed, such as generating new data, either from scratch~\citep{mishra2024mathcamps} or by using existing datasets as seeds~\citep{zhou2024your}, it is not clear whether these can scale to more sophisticated mathematics, as the methods were tested using mathematics on the level of grade-school, and it is unclear how well they would scale to much higher levels of mathematical abstractions and problem-solving difficulty (the two main metrics of difficulty, as outlined in~\ref{subsubsec:difficulty}.

\paragraph{Recommendations.} If the dataset is sufficiently large, it is advisable to keep a part of the dataset hidden from the public so that the dataset creators can compare how well newly-released, publicly accessible open-weight models score in the public versus the hidden part of the dataset. Diverging scores on two splits of the dataset can indicate contamination. Care must be taken that datapoints from the hidden datasets are from the same distribution compared. For mathematics, this means at minimum: Same domain, difficulty range, and proof technique. This can be challenging: Given a specific problem $P$, it takes effort to source a new problem $P^{\star}$, that on all relevant metrics (including the mentioned ones) is similar to $P$, but at the same time is not completely analogous to $P$ -- as it would be by, e.g., merely changing items in the problem statement that have a negligible on the proof (this is an approach followed by GSM8K-Symbolic dataset~\citep{mirzadeh2024gsm}, which nonetheless turned out to be challenging for LLMs, highlighting their current limited reasoning capabilities). The approach advocated here has limitations for models released as-a-service~\citep{la2024language}, via APIs or GUIs, as running the hidden dataset through risks it automatically being used at a later stage as training data for those models. RealMath proposes a continuously refreshed benchmark mined from arXiv papers and forum discussions, which reduces the half-life of static public test sets~\citep{realmath2025}.

\subsubsection{Binned Sankey diagrams}
\label{subsec:binned-sankey}

Math benchmarks are often reported with aggregate statistics such as accuracy, pass@$k$, or majority-vote accuracy. These statistics compress item-level outcomes into a small number of scalars. On mixed benchmarks, an average gain can coincide with regressions on specific slices (topic, modality, or answer format), which the average does not show. When assessing model improvement over time, this information conveys both too little detail and is also hard to read.
This issue is not solely applicable to mathematics, but is very prevalent in this case.

\paragraph{Recommendation.} To support better visualizations, we recommend the use of binned Sankey diagrams. These are item-level transition plots between two evaluations. Each benchmark item is assigned to an outcome bin under a baseline evaluation (for example, model $M_t$) and to an outcome bin under a second evaluation (model $M_{t+1}$). The diagram draws flows from baseline bins to new bins, with band widths proportional to the number of items making each transition. Sankey-style migration plots have been used in other evaluation contexts; see Fig.~2 in \cite{frieder2023mathematical}.

For mathematics, the visualization works best in the following way: Choose two evaluation points and define a small set of outcome bins. The minimal binning is binary (\texttt{correct}/\texttt{incorrect}). If annotations exist, bins can reflect error types~\citep{frieder2023mathematical} (forsexample, algebra error, invalid step, wrong final answer) or partial credit (for example, \texttt{solved}, \texttt{nearly solved}, \texttt{off-track}). If the evaluation involves tools, bins can separate \texttt{solved without tools}, \texttt{solved with tools}, and \texttt{failed}. The resulting diagram makes regressions explicit. A flow from \texttt{solved} to \texttt{failed} remains visible even when net accuracy increases. For heterogeneous benchmarks, the same transition plot can be computed within slices such as topic, difficulty level, or representation (symbolic, word problem, diagram-based, or formal proof assistant). Slice-level transitions show whether gains are concentrated in a subset of items and whether any slice exhibits systematic regressions. 

Benchmark reports can include a binned Sankey diagram alongside aggregate metrics when comparing model versions or training regimes. This requires storing per-item outcomes (and, when applicable, slice labels) so that transitions can be reproduced. Using the same bins across releases supports longitudinal comparisons.

\subsection{Distinct Issues}
This section will detail issues that specifically appear with either natural language or formal language datasets, or mixed datasets. 
The most prominent difficulty with natural language datasets is the lack of trusted automatic evaluation as will be explained in Section~\ref{subsec: auto eval}.

A natural way to circumvent the difficulty of evaluating natural language proofs is by making use of formal language datasets. Instead of having an LLM generate natural language mathematical proofs, one can train it to generate proofs in a verifiable formal language such as Lean or Isabelle. 
Therefore, existing libraries of formal proofs can play a crucial role in the development of machine learning-based automatic theorem provers. Sections~\ref{subsec: tool misalignment}~and~\ref{subsec: data duplication} will provide details on how some of the largest existing collections of formal proofs have been and are being created. This is followed by an analysis of how this process affects the use of formal proof libraries as evaluation datasets; with some remarks relevant to using formal libraries for training.
More concretely, Isabelle's Archive of Formal Proofs (AFP) and Lean's Mathlib will be investigated. This choice is due to the authors' familiarity with these libraries but many of the points mentioned below will also be valid (at least to some extent) for other systems.
One particular issue relevant to all datasets is their scale. The lack of a large-scale dataset consisting of pairs of formal and informal mathematics is a significant bottleneck for autoformalization. Current datasets, such as miniF2F are on the order of a few hundred datapoints, which are enough for assessing autoformalization, but not for training models to support autoformalization. A recent effort~\citep{ying2024lean} scales this to 57k datapoints of pairs. Nonetheless, this scale is not yet comparable to the scale at which LLMs can be trained on informal mathematics. 

\subsubsection{Lack of Trusted Automatic Evaluation}
\label{subsec: auto eval}

The easiest way to automate the evaluation of mathematical prompts is to formulate them in a way such that the answer can be represented as a single token (e.g., a number or a mathematical term). Automatic evaluation can then be performed by keyword-matching the token to the gold-truth answer. A slightly more advanced variant is pursued by the MATH dataset, which allows a reasoning section, but encloses the final answer in a \texttt{\textbackslash boxed} environment (and only this is keyword-matched).

No large machine learning dataset on mathematics currently exists that allows an arbitrary proof of an AI system to be checked for correctness against the gold-standard proof in the evaluation dataset. The Open Proof Corpus collects expert judgments on model-generated natural-language proofs and reports large gaps between boxed-answer accuracy and proof validity~\citep{openproofcorpus2025}. A proof-of-concept, highlighted by the IMO Small Challenge\footnote{\url{www.imo-small-challenge.io}}, shows how partial automation might be achieved, where a necessary test for correctness is carried out using detailed proof annotation. This builds on the idea that a system like an LLM, combined with a deterministic form of matching, can at least exclude incorrect proof candidates. If the ground truth is sufficiently rich, the annotations help the LLM break down a proof and understand its main features (accepting that subtler points of proofs may still be currently out of reach of LLMs' understanding).

Some success has been attained in using an LLM to grade (or teach) another LLM~\citep{eldan2023tinystories, mukherjee2023orca}. Yet, for mathematics, we believe that this is not necessarily the right approach forward. The current generation of models does not have a sufficiently high performance on mathematics to be used as graders: for tasks that would often be deemed simpler than mathematics, this approach can fail~\citep{wang2023large}, although for solving programming puzzles this approach was successful~\citep{haluptzok2022language}. Anecdotal evidence suggests that adapting the approach to proofs is challenging because of the multitude of potential proof variations for a single statement, which gives rise to a diverse set of textual expressions.
At present, grading can, at best, be automated by the use of detailed human annotation for each problem in combination with LLM assistance based on such annotation. 

\paragraph{Recommendations.} In general, we recommend developing evaluation methodologies that allow for natural language proofs to be checked for correctness. This is difficult in general. A middle ground is to include as much metadata as possible when annotating datasets with manual effort, with a view towards supporting LLMs in using this metadata to assess proof candidates.
Second, we recommend more research that quantifies the degree to which LLMs can evaluate mathematics. This will help in making objective statements about how good or how poor LLMs are at evaluating different kinds of mathematics, and help in identifying areas for improvement. 

\subsubsection{Tool Misalignment} \label{subsec: tool misalignment}

As writing formal proofs imposes slightly different challenges to writing informal proofs, it is important to distinguish mathematical proficiency from proficiency in using tools provided by the formal environment. 

One cause of this discrepancy is that side conditions deemed trivial in conventional mathematical literature have to be formalized as diligently as the rest of the mathematical text. On the other hand, proving many statements considered routine by mathematicians can be mechanized. 
For example, proving that a function is continuous often corresponds to choosing the correct subset of a few relevant lemmas and this can be achieved with a simple search algorithm, such as Lean's continuity tactic.
Next to this domain-specific formalization aid, there are also general-purpose automation tools that can be used to formalize certain simple statements. For example, Isabelle's Sledgehammer tool~\citep{sledgehammer2007, sledgehammer2008} translates the given statement so that it can be understood by a number of external automatic theorem provers (ATPs). The ATPs, in turn, attempt to prove the statement and send the proof back to Isabelle. 

\paragraph{Automation and Proofs}

Naturally, the existence of automation affects how the task of formalization is approached: Without automation, formalization would consist of translating a mathematical proof step-by-step into the ITP's logic. But having access to advanced tactics and general-purpose ATPs, one only needs to repeatedly break down the proof into smaller pieces until these can be tackled by automation. 

Eventually, this can lead to proofs being formalized in a way that is not very close to the original natural language proof. For example, elementary homework-style problems often consist of long calculations that might correspond to one single tactic invocation in the formal system. Also, in more advanced topics, the automation of a theorem prover can make the proof less readable or skew its focus. For example, natural language proofs typically make it very clear which definitions need to be unfolded and when. In Lean, such unfolding does not have to be made explicit, which can widen the gap between formal and informal. 

The impact of these mechanisms needs to be kept in mind when formal datasets such as Lean's Mathlib or Isabelle's AFP are used for LLM evaluation. In particular, it is likely that current datasets overly reward models for their ability to deploy the language-specific automation, rather than complete ``understanding" of the underlying mathematics.
As some evidence of this, \citet{hu2024minictxneuraltheoremproving} show that LLM performance drops significantly when common automation tactics are turned off, even for theorems that have human-written proofs that do not rely on the automated tactics.

\paragraph{Strong automation and the Archive of Formal Proofs (AFP)}

These matters are especially significant with the Isabelle AFP because of the powerful general-purpose automation that has been widely used in its development over the last decade. It can therefore be expected that a large subset of theorems in today's AFP have been formalized by repeatedly breaking them down into smaller subproblems until Sledgehammer could find their proof. 

A natural way to turn the AFP into a proof dataset is by splitting up all formal proofs into their individual steps~\citep{jiang2021lisa}. However, by the above characterization, such a dataset might be skewed towards being tractable for automation because of the particular way in which AFP theorems have come into existence. This can make it hard to assess the level of difficulty in the problem set. 

Especially when equipping LLMs with the ability to access existing automation, careful evaluation is essential. It is important to determine whether the model genuinely \enquote{understands} complex mathematical concepts and can scale this understanding up to more difficult problems; the alternative being that the model merely excels at finding \enquote{low-hanging fruits}, which, when combined with non-ML automation, might appear deceptively impressive.

We would like to stress that we believe that the use of formal mathematics datasets such as the AFP is a valid evaluation tool. The results of recent publications~~\citep{first2023baldur, thor, magnushammer} are especially promising, and we are looking forward to their continuation. Nevertheless, we believe that future evaluation should carefully consider the impact of existing automation in ITPs on their libraries. 

\paragraph{Recommendations.} The available formal mathematical libraries that include deep, research-level results have generally been created using a significant amount of automation. Researchers could create small, specific additional test sets of proofs that are formalized without automation and evaluate mathematical assistants on these. 

\subsubsection{Non-Trivial Data Duplication} \label{subsec: data duplication}

Next to these aspects related to the nature of theorem provers, there are also challenges intrinsic to mathematics itself. In particular, mathematics has a self-similar structure: there are profound connections between areas that are seemingly disconnected, and many mathematical statements have several related versions, depending on the perspective and level of generality chosen. This section will detail some of the inherent challenges this creates for the construction of formal mathematical libraries and then consider the impact of using them as datasets. 

Ideally, a formal library should have as little duplication as possible to allow for a smoother user experience and to reduce the amount of code maintenance required. At the same time, this ideal can seldom be upheld fully since formalizing theorems only in their most general form is intractable. Some level of non-trivial duplication is therefore accepted and common in formalization. 

As an example, Lean's Mathlib often provides several versions of a lemma,\footnote{We note that this facility is not exclusive to Lean; e.g., Isabelle provides several versions of a lemma.} which makes it more convenient to use the library. Instead of having to ``import'' a lemma and then transform it into the right format, one can directly refer to commonly used variants. For example, the statement that 
\begin{equation}
    a + b + (c + d) = a + c + (b + d)
\end{equation}
in a commutative semigroup is explicitly formalized even though it could easily be derived from associativity and commutativity whenever needed. Furthermore, there is support for automatically generating lemmas, e.g., to generate the additive version of an abstract algebra statement given in terms of multiplication. 

Next to this, the rules and customs of how a collection of formal proofs is built and maintained have a significant impact on its properties as a machine-learning dataset. Compared with Mathlib, the Isabelle AFP is more static and aims at being archival, i.e.,  providing long-lasting support for its content and not removing previously defined concepts. This makes duplication inevitable as soon as parts of the library go through a larger restructuring or even redevelopment. For example, there are two separate algebra libraries for Isabelle, one in the AFP and a second one using a different formalization approach in the Isabelle HOL-Library, which often gets used in conjunction with the AFP. 

As a further example, consider the AFP section on category theory. Currently, there are at least five separate formalizations of category theory at various stages of development with all of them taking different approaches\footnote{cf.~\url{https://www.isa-afp.org/topics/mathematics/category-theory/} not counting generalizations of category theory and continuations of previous developments.}. This is likely to be for historical reasons, but also because exploring different representations of category theory is an interesting topic in mathematical foundations. 

Naturally, if duplication is present in the formal library, a lot of care becomes necessary when designing machine-learning datasets from it. If data is just randomly split into a training and validation set, then leakage can occur on one hand. Of course, the duplication will typically be on the level of mathematical concepts and not exist as a verbatim doubling of formal language code. Nevertheless, disregarding the issue of duplication makes it hard to gauge to what extent a model manages to produce formal proofs from the ground up. Possibly, parts of the model's performance simply stem from learning how to translate between different representations of the same mathematical concept. On the other hand, duplication brings to the forefront the issue of data representation, as machine learning models will only perform well on that representation on which they have been trained, which is in contrast to human mathematical reasoning, which, to a degree, is robust against different formal representations of the same mathematical objects.

Translation of formal representations constitutes a highly interesting and relevant skill in its own right. Nevertheless, good test design should enable researchers to judge to which degree the respective skills are achieved.

\paragraph{Recommendations.} 
Following the discussion above, we recommend developing methodologies that ensure that train/test splits adequately measure a model's ability to generalize. One direction is to test on new formalizations that occur after a model has been trained, as was explored in \cite{hu2024minictxneuraltheoremproving}. Nevertheless, care should be taken to ensure that the mathematics being formalized does not occur in previous projects. Hence, we particularly recommend evaluating on domains that have not previously been formalized.

More broadly, the ideal evaluation of digital mathematical assistants should include case studies in which previously unformalized proofs get formalized using the assistant. There are various ways to realize this. On a small scale, researchers can qualitatively evaluate the LLM by formalizing some reasonably difficult, previously unformalized proof. On a larger scale, the assistant can be provided as a plugin to theorem provers, which a large number of people working in formalization can then use in real-world scenarios. If there are enough users, such a setup could even use A/B testing for multiple architectures of mathematical assistants.

\subsubsection{Version Fragility and Tactic Stability}
Formal proof corpora are often treated as static datasets consisting of a theorem statement plus a proof script. In practice, they are tightly coupled to a moving toolchain (involving a specific version of the prover, a particular commit of the library, and details of automation implied by this combination). This can make them fairly brittle as datasets: even small library changes can invalidate proofs or silently change what intermediate tactics do.

A common failure mode is ``tactic instability'', where a proof uses a flexible tactic (in Lean, most notably \texttt{simp}) and then applies a subsequent tactic whose success depends on the exact transformed goal. Because the set of parameters used by the automated tactic (e.g., lemmas used by \texttt{simp}) evolves, the transformed goal may drift over time, making the downstream step brittle even if the overall theorem statement is unchanged.

\paragraph{Recommendations.}
Recent work in Mathlib\footnote{See \url{https://leanprover-community.github.io/mathlib4_docs/Mathlib/Tactic/Linter/FlexibleLinter.html} for details.} introduces a stability linter (``FlexibleLinter'') that flags precisely these patterns, encouraging authors to make tactic behavior more explicit (e.g., by specifying a minimal set of lemmas used in the simplification). As a dataset design principle, proofs that pass such stability linting are better candidates for long-lived benchmarks, since their execution is less sensitive to future library drift.

\subsubsection{Hidden Context Dependence: Local Instances and Attribute State}
Even in a fully formal setting, a dataset item is seldom determined by the goal alone. In Lean, imports, local instances, and attribute state can change the meaning of a line of proof text. This matters for dataset construction because ``the same'' proof script can behave differently when replayed under a different environment.

\paragraph{Example (instance ``pollution'').}
In measure-theoretic developments in Lean, the proposition \linebreak \texttt{MeasurableSet s} is parameterized by an implicit measurable-space instance. If a proof introduces a local measurable space \texttt{m0} but an ambient instance \texttt{mW} remains in scope, \texttt{MeasurableSet} may silently refer to \texttt{mW} unless the instance is pinned:

\begin{Verbatim}[fontsize=\small,frame=single]
-- Fails: `MeasurableSet` uses the ambient instance, not the local one
have h_bad : MeasurableSet s := by
  simpa using hs

-- Works: pin the intended instance explicitly
have h_ok : @MeasurableSet alpha m0 s := by
  simpa [m0] using hs
\end{Verbatim}

This illustrate that storing only the final proof term (or only the local goal) in a dataset can be insufficient to replay a proof deterministically, because replay depends on the surrounding instance environment and imported modules. For more examples from various areas of math, see the reference patterns contained in the Claude Code skill for Lean 4 \citep{FreerLean4Skills}.

\paragraph{Example (context-dependent simplification).}
Automation can also be context-sensitive in a more mundane way. Automated tactics such as an instance of \texttt{simp} in Lean can succeed or fail depending on hypotheses introduced earlier in the proof:

\begin{Verbatim}[fontsize=\small,frame=single]
-- Works with extra context in scope
filter_upwards [hw, hu_eps, hv_eps] with w hu hv
simp [Real.norm_eq_abs]

-- May fail without that context, even if the syntactic goal looks similar
simp [Real.norm_eq_abs]
\end{Verbatim}

This suggests that proof snippets that are extracted without their surrounding context (imports, local hypotheses, attributes) can be underspecified as dataset items for training or evaluation.

\subsubsection{Resource Sensitivity: Heartbeats and Elaboration as Part of the Task}
In interactive theorem provers, ``correctness'' is not clearly delineated from performance.  A proof that compiles only under extremely permissive resource limits (such as ``heartbeats'', recursion depth, or wall-clock time) is often not usable in real formalization workflows, and may not be reproducible across environments (especially if the high resource bounds are not clearly specified).

Lean exposes heartbeat and other resource limits directly: 
\begin{Verbatim}[fontsize=\small,frame=single]
set_option maxHeartbeats 40000
set_option synthInstance.maxHeartbeats 40000
set_option maxRecDepth 2000
\end{Verbatim}

These resource bounds can be exceeded by complex elaborations or typeclass searches. Sometimes this is due to the complexity of the underlying mathematics, but other times it is from ``fighting the elaborator''.  For instance, leaving instance arguments implicit can trigger expensive unification/typeclass search, sometimes consuming hundreds of thousands of heartbeats before timing out.  This suggests that datasets and benchmarks should treat resource budgets as first-class metadata, and report resource usage alongside whether the code passed or failed.

\subsubsection{Redundant Scaffolding and Accidental Complexity in Proof Scripts}
Formal corpora often contain a lot of proof-script structure that is not mathematically essential, including redundant tactic steps, arguments to automation that are not used, and lemma variants that exist only to satisfy local engineering constraints.
As dataset artifacts, these patterns can dominate the training signal while teaching models brittle or unidiomatic behavior.

\paragraph{Example (no-op tactic after automation).}
A common failure mode when replaying mined proof scripts is that one tactic already closes the goal, making subsequent steps ill-typed. For example, in Lean, this sequence can fail:

\begin{Verbatim}[fontsize=\small,frame=single]
simp  -- closes the goal
aesop -- error: no goals to be solved
\end{Verbatim}

\paragraph{Example (lintable redundancy in automation arguments).}
Even when scripts compile, they often include extra automation arguments that are unused in practice.  Mathlib-style linting can flag such redundancy and enable construction of a cleaned corpus:

\begin{Verbatim}[fontsize=\small,frame=single]
-- Linter: unused simp arguments (suggest removing to reduce clutter)
simp [pow_two, mul_assoc, mul_add, mul_comm, add_comm, add_left_comm]
\end{Verbatim}

In light of this, it may be useful for datasets to include both a raw corpus (for realism) and a linter-cleaned corpus (for learning canonical structure).

\section{The Future -- Novel Datasets to Support Mathematical Copilots}
\label{sec: future}

In this section, we cover various aspects of mathematical research practice that, to date, are not covered at all by any of the existing datasets. This contrasts with the previous section on mathematical datasets that covered parts of mathematical practice, albeit with the noted limitations. Current datasets are exclusively focused on publishing the \emph{results} rather than the intermediate steps that a mathematician goes through in the process of devising a proof. The different types of workflows, the proof-counterexample dialectic, and proof transfer,
etc., are all examples of intermediate processes that do not have clear representations in data that can be used to fine-tune LLMs. Contrasting with the previous section, where we made specific recommendations on how to improve \emph{existing} datasets, in this section, we do not make specific recommendations, as our recommendation is to simply devise such datasets. While a few inroads have been made into non-proof-based datasets, see Section~\ref{sec:standardized} these are still very much at their inception, and, to our knowledge, the GHOSTS dataset is the only attempt to date to try to \enquote{data-ify} some aspects (solely related to mathematical literature search) of daily mathematical process.

An example of proof transfer is given by the Ax--Grothendieck theorem, which states that every injective polynomial map $\mathbb{C}^n \to \mathbb{C}^n$ is surjective; rather than proving this directly over $\mathbb{C}$, one observes that the statement is trivial for finite fields -- any injective map on a finite set is bijective -- and then uses model-theoretic compactness to transfer the result to algebraically closed fields of characteristic zero. Here, one does not redo the argument in the new structure directly; instead, one uses model theory to move the truth of a sentence from one place to another.

\subsection{Mapping Mathematical Workflows to Data}
\label{subsec: workflows}

The previous sections have shown that existing datasets and benchmarks only deal with specific mathematical tasks. In particular, they focus mostly on question answering and theorem proving, with solutions being presented in a streamlined way (as is typical for results in mathematical textbooks). In consequence, they do not sufficiently cover all steps in typical mathematical workflows, such as surveying mathematical topics, gathering related results, establishing high-level proof strategies and intuition, refactoring proofs, or carrying out field-specific routines. Yet, we would like to evaluate and train mathematical copilots on these tasks, too, in order to obtain full-spectrum assistance across all facets of mathematical research practice.
We note that datasets used to pre-train LLMs exhibit broader coverage, e.g., of mathematics-related Q\&A communities, blogs, and educational material; however, it is hard to assess and control the quality of the included mathematical content. This results in a weak learning signal and, for a strong mathematical copilot, one likely requires fine-tuning on a comprehensive, high-quality dataset of mathematical workflow steps.

\paragraph{Simple methods as precursors.}
Even relatively simple machine learning approaches have produced genuine mathematical discoveries.
Wagner~\citep{Wagner:2021Constructing} demonstrated that reinforcement learning, framed as a single-player graph-building game, can find counterexamples to open conjectures in extremal combinatorics.
PatternBoost~\citep{Charton:2024Patternboost}, built on Karpathy's character-level language model makemore~\citep{karpathy2022makemore}, alternates between local search and transformer-based generation to discover constructions that refuted a 30-year-old conjecture.
These methods succeed precisely because they operate in domains where candidate solutions can be automatically evaluated. This is a form of verifiable feedback that sidesteps the hallucination problems plaguing open-ended generation.
They are precursors to more general workflows that combine neural generation with symbolic verification and human mathematical insight.

We advocate a more bottom-up approach that consists of creating a taxonomy of workflows, i.e., isolating and categorizing workflow steps.
Specifically, many mathematical workflows can be modularized into smaller sequential steps, which themselves can be ordered along different dimensions, e.g., their mathematical depth, level of abstraction (from general, high-level proof techniques and principles to specialized, problem-specific approaches), mathematical subject, how well a symbolic approach might handle them, whether the workflow step is strictly mathematical or meta-mathematical (e.g., literature search). With this in view, we distinguish between general \emph{global} workflows, which are used across mathematical fields, and field-specific \emph{local} workflows.  For instance, we refer to the \emph{Tricki}\footnote{\url{https://www.tricki.org}} for a collection of problem-solving techniques that can be viewed as abstract workflows. General proof techniques, such as linearization and fixed point theory, can be viewed as global workflows. However, we note that specific instantiations of such techniques, e.g., related to dynamical systems, can also be classified as local workflows. Examples of even more local workflows include approaches tailored to specific kinds of partial differential equations, e.g., the method of characteristics, maximum principle, energy estimates, or Green's functions, as these methods frequently occur as individual steps in larger chains of arguments in research-level mathematics. 

There is no hard boundary for transitioning between local and global workflows. For example, the \enquote{routine} task in analysis to upper bound an expression efficiently, as opposed to evaluating it exactly, is found both in elementary contexts, such as proving inequalities such as versions of arithmetic mean-geometric mean inequality, up to more advanced cases, such as for ordinary differential equations (ODEs) or partial differential equations (PDEs), where a routine task is that of applying an \enquote{energy method} or \enquote{Lyapunov functional method}. We, therefore, deliberately do not draw a specific boundary of when to categorize a workflow step as local vs. global.

While creating a complete taxonomy of workflow steps is beyond the scope of this article, we want to examine selected workflows and their steps (both local and global) in order to highlight issues in translating these to data and specific challenges. We note that these are merely illustrative examples meant to show how first steps could be undertaken to generate datasets that support such workflows.

While a workflow (indicated with arrows below) can, in principle, easily be mapped to data by converting it to $n$ tuples of datapoints (where $n$ denotes the number of workflow steps) and filling text in between, in practice, in certain cases, issues may arise. We give examples of both global and local workflows below and highlight both examples where these can be transformed to data in a straightforward manner, as well as cases in which this transformation is more difficult. We urge the community to explore such translation of workflows to data further.
Future work should address how such workflow datasets might be structured, how symbolic or numeric methods can be integrated into individual workflow steps, and which parts of these workflows are best left to humans versus delegated to a mathematical copilot. Once such datasets exist, LLMs can be trained on them using standard methods such as expert iteration~\citep{DBLP:conf/nips/AnthonyTB17,havrilla2024teachinglargelanguagemodels}.

\paragraph{Inference-time compute in reinforcement learning needs problem analyses.}
Several recent math reasoning systems use substantial inference-time compute, for example, by sampling many candidate solutions, running search over intermediate steps, or iterating tool calls, and then using reinforcement learning to train policies that decide which branches to expand.
In domains with automatic verification, these loops can rely on a binary reward signal.
For most informal mathematical tasks, that signal is too sparse: a model receives feedback only after a long chain of steps, and failure provides little information about where the attempt went off track.

One remedy is to include, alongside each problem statement, a structured \emph{problem analysis} that makes intermediate progress measurable.
Concretely, this can include a normalized restatement of the problem, the relevant definitions and constraints, a decomposition into subgoals that a solver would aim to establish, and examples of admissible intermediate claims (or failure modes) that an evaluator can recognize.
These annotations can be authored by experts or drafted with model assistance and then checked, but they are rarely present in current math benchmarks.
Without this layer, RL methods that spend compute at inference time are pushed toward brute-force sampling on a sparse objective, which does not scale to longer proofs or to tasks whose correctness is not cheaply checkable.

\paragraph{Global workflows}    
Literature search is one of the most general workflows across the sciences, having particularities in how it takes place in mathematics. Prior work can be related to the problem a mathematician has at hand in many ways. One might, for instance, want to know what is known about a particular mathematical object at hand. The context of a problem might have many options for what to look for, so the first step in the workflow might be to identify an object that has a good chance of having been seen before. One example is integer sequences: perhaps the solutions to a combinatorial problem are computable in small cases, and it is often useful to know if the resulting sequence has been encountered in the context of other problems (there might be many ways to get a sequence: for instance, we might have to choose variables to fix and then only vary one parameter of the problem). For this particular case, the Online Encyclopedia of Integer Sequences\footnote{\url{https://oeis.org/}} is a well-known, efficient, specialized tool, with many pointers to the literature, and the ability to recognize partial matches (e.g., perhaps one's sequence is coarser than the one from relevant prior work). In case there are no matches, we can try to repeat the process with other candidate objects; in case there are, the challenge turns to assess whether any of the results seem meaningfully related to the current case (and given only a few integers, there are often too many sequences containing them). For instance, the sequence $1, 4, 44$ appears in path-counting problems in some lattices, or in a few number-theoretic contexts (like products of odd-indexed Lucas numbers), as well as many other contexts, and more context that was not included during the search might be relevant in identifying which of these sequences might be meaningfully related to the current problem. We could broadly summarize this workflow as:

     \emph{Identify objects to search for $\rightarrow$ find related work describing the object $\rightarrow$ assess potential relation to current context}

We might expect LLMs to help generalize this process beyond what specialized tools, such as the OEIS, are capable of. Most mathematical objects are significantly harder to describe than integer sequences: for instance, one might have a particular topology on a function space that might be unique to the current context, but something of the same ``shape'' might have been seen before. The assistant would have to recognize and know how to describe this shape, and identify matches across potential differences in what definitions are implicitly or explicitly used, as well as potential equivalent definitions that might cause a description in the literature to differ on the surface (e.g., mention ``accumulation points'' vs ``limit point'').

Examples of this workflow for training and evaluation might be automatically extractable in a post-hoc fashion from the mathematical literature itself: one might be able to look at how previous work is referenced in existing proofs as a source of examples of relevant previous work (other matches for the same object that are not the paper that was cited are most likely examples of spurious matches).

\paragraph{Local workflows} 
We list below examples from several distinct subfields of mathematics (differential equations, knot theory/low-dimensional topology, and number theory).
\begin{itemize}
    \item Suppose, for instance, one has a first-order ODE  $\partial_t u = F(u)$ with some initial data $u(0) = u_0$ and wants to know how the solution grows in time.  A standard technique is to introduce a key functional $E(u)$ of the solution (often something like an \enquote{energy} or \enquote{Lyapunov functional}) and then compute the derivative $\partial_t E(u)$ using the ODE and the chain rule (for PDEs, one often has to also perform several times integration by parts).  Then, one bounds this derivative as best one can.
The goal is to reach some differential inequality of Gronwall type, e.g., $\partial_t E(u) \leq C E(u)$, so that a Gronwall-type lemma may be applied (but it is not always precisely the Gronwall lemma, but something similar). 

Summarizing, the workflow is thus of the form:
\begin{center}
     \emph{ODE/PDE $\rightarrow$ find functional $\rightarrow$ compute derivative of functional $\rightarrow$ bound derivative $\rightarrow$ reach Gronwall-type inequality}
\end{center}
We note that individual steps of this may be solved using symbolic-numeric methods, which opens the possibility of using a tool-integrated-reasoning (TIR) approach to enable an LLM to discharge these proof steps to symbolic or numeric tools to, e.g., compute derivatives symbolically. Thus, one option to encapsulate these workflow steps in data is by using datapoints whose form is
\begin{center}
     \emph{(ODE/PDE, functional, functional derivative, derivative bound, Gronwall-type inequality)}
\end{center}
with potentially filler text between the raw mathematical objects contained in the datapoint (i.e., the ODE, the functional, etc.). The raw mathematical objects should be symbolically or numerically generated so that custom filler text can be added in between them, and an LLM can observe a range of ways in which to speak and textually connect a fixed tuple of mathematical objects.

What one typically wants, as a mathematician, in the case of this workflow, is to try various guesses for the functional. If a TIR approach is used, this data representation of this workflow should specifically offer \enquote{entry points} to make it easy for an LLM to accept arbitrary input at certain places and use tools to automatically run the essential parts of the workflow. An LLM is then much better equipped to run the workflow steps and produce an estimate, either automatically or semi-automatically. One can preliminarily do this already through conversation with current state-of-the-art LLMs, but with a lot of mistakes on the LLM's part, which a stronger grounding in data would help to correct.

\item A commonly encountered workflow for evolutionary PDEs (i.e., PDEs whose solution evolves with a time variable $t$), analogous to the first-order ODE example given above, can be illustrated as follows. One wants to find a functional which will facilitate providing useful information about solutions of the PDE. To illustrate how such functionals are found in practice, we begin by considering the simple example of the linear wave equation $\partial_{tt} u - \Delta u = 0$ on a bounded domain $\Omega \subset \mathbb{R}^n$, with smooth boundary $\partial \Omega$, subject to homogeneous Dirichlet boundary conditions $u=0$ on $\partial \Omega$ and initial conditions $u(x,0)=u_0(x)$, $\partial_t u(x,0)=u_1(x)$. (One can also study this problem on an unbounded domain $\Omega$ such as, for instance, $\mathbb R^n$, and impose that solutions must decay in the unbounded directions of $\Omega$.) To show energy conservation, we define the energy functional 
\begin{equation}\label{eq:definition_of_energy}
    E(t) = \frac{1}{2} \int_{\Omega} \Bigl[ (\partial_t u)^2 + |\nabla u|^2 \Bigr] \, dx,
\end{equation}
which represents the sum of kinetic energy and potential energy. Differentiating $E(t)$ in time under the integral sign, we obtain 
\[
\frac{dE}{dt} 
= \int_{\Omega} (\partial_t u)\,(\partial_{tt} u)\, dx 
+ \int_{\Omega} \nabla u \cdot \nabla(\partial_t u)\, dx.
\]
An integration by parts shifts the spatial derivative in the second term, while the boundary contribution vanishes because $u$ and hence $\partial_t u$ are zero on $\partial \Omega$. By substituting the wave equation $\partial_{tt} u = \Delta u$, the expression collapses to zero, showing that $\tfrac{dE}{dt} = 0$. Thus 
\begin{equation}\label{eq:conservation_of_energy}
    E(t) = E(0) \text{ for all }t>0.
\end{equation}

More generally, on any finite time interval $[0,T]$, one can often prove an energy bound of the form
$E(t)\le C(T)\,E(0)$ for $0\le t\le T$, with $C(T)$ possibly being a constant function. When $C(T) = C>0$ is constant, we write
\begin{equation}\label{eq:uniformenergybounds}
    E(t) \le C E(0) \text{ for all }t>0.
\end{equation} An inequality like \eqref{eq:uniformenergybounds} is sometimes referred to as an \emph{energy estimate}. Notice that the conservative situation \eqref{eq:conservation_of_energy} satisfies the inequality \eqref{eq:uniformenergybounds} with $C=1$. 
The energy bound \eqref{eq:uniformenergybounds} can be proved for certain wave equations with non-constant coefficients and for some non-linearities (often under additional structural or smallness assumptions); it provides useful control on the solution. For example, it can be used to prove the uniqueness of solutions of the wave equation \cite[Section 2.4, Theorem 5]{evans2010pde}, as well as finite propagation speed for solutions of the (constant-coefficient) wave equation \cite[Section 2.4, Theorem 6]{evans2010pde}, and more generally under standard hyperbolicity assumptions.

One can invert the previous perspective on this problem:  if one had no prior knowledge of this equation and its properties, how could equations \eqref{eq:conservation_of_energy} or \eqref{eq:uniformenergybounds} be derived in the first place? In particular, to obtain an \emph{energy identity} (and hence an energy estimate) for this model problem, one takes the equation $\partial_{tt}u - \Delta u=0$ and multiplies both sides of it by $\partial_t u$, and then integrates by parts on $[0,T] \times \Omega$, where $T>0$. (The exact choice of this multiplier depends on the equation at hand; as we will see in the next paragraph, $u$ turns out to be a good multiplier for the heat equation. Moreover, multiple good choices of multipliers are usually known to exist for each equation, with different multipliers usually leading to different and useful inequalities, among which \eqref{eq:uniformenergybounds} can be viewed as one of the simplest examples.) Assuming that either $u=0$ on $\partial\Omega$ (Dirichlet, hence $\partial_t u=0$ on $\partial\Omega$)
or $\partial_\nu u=0$ on $\partial\Omega$ (Neumann), after a few integrations by parts, one obtains exactly that the energy functional defined in \eqref{eq:definition_of_energy} is conserved: $E(t)=E(0)$ for all $t \ge 0$. It turns out that for many more general wave equations of the form $Pu = F$ (where $P$ is a wave-type differential operator, e.g.\ $\partial_{tt}-\Delta$ with variable coefficients and lower-order terms, and $F$ is a forcing term), the same choice of multiplier $\partial_t u$ (meaning we multiply the equation by $\partial_t u$ and integrate by parts) yields an energy identity or inequality. Under additional structural assumptions, this can be turned into a uniform-in-time bound of the form \eqref{eq:uniformenergybounds}. In practice, finding good multiplier functions and performing integration by parts in the resulting manipulations is one of the standard ways new inequalities are discovered.

Let us modify the wave equation argument slightly to obtain an analogous result for the linear heat equation $\partial_t u - \Delta u = 0$. In this case, as we will now see, one obtains
\begin{equation}\label{eq:heat_equation_inequality}
    E(t) \le E(0), \qquad t \geq 0
\end{equation}
where $E(t)$ is defined to be $E(t) = \frac{1}{2} \int_{\Omega} u^2 \, dx$; notice that this functional $E$ differs from the functional we used for the wave equation. As in the wave equation case, the inequality \eqref{eq:heat_equation_inequality} can be discovered by multiplying both sides of the heat equation by an appropriate function (here, $u$), and then integrating on the domain $[0,T]\times \Omega$. After a few uses of integration by parts, if $u=0$ on $\partial\Omega$ (Dirichlet) or $\partial_\nu u=0$, then one obtains, for $0 \le t \le T$,
\[
E(t) + \int_0^t \|\nabla u(t,\cdot)\|_{L^2(\Omega)}^2 \, dt = E(0).
\]
Since $\int_0^t \|\nabla u(t,\cdot)\|_{L^2(\Omega)}^2 \, dt \geq 0$, one concludes \eqref{eq:heat_equation_inequality}. The inequality \eqref{eq:heat_equation_inequality} can be used to prove the uniqueness of solutions to the heat equation; one can find in \cite[Section 2.3, Theorem 10]{evans2010pde} a proof of uniqueness of solutions forward in time, and \cite[Section 2.3, Theorem 11]{evans2010pde} proves a backward uniqueness property. (Unlike the wave equation, the notions of uniqueness forward and backwards in time do not ``coincide'' immediately for the heat equation.) The idea of manipulating a useful functional such as $E(t)$ goes by the name of ``energy methods'' in the PDE literature. The ``energy method'' for the heat equation serves as a template in the following sense: many more difficult non-linear PDEs (of which the Navier--Stokes system is one example) admit a version of this ``multiply by the unknown and integrate'' trick, and understanding it in the heat equation model helps one to understand those.

The main insight in these ``energy method'' arguments for both equations was the introduction of the energy functional $E(t)$. 
We saw that for both the heat and wave equations, this functional can be derived by multiplying the equation with an appropriately chosen multiplier---specifically, $u$ for the heat equation and  $\partial_t u$ for the wave equation---and then integrating by parts over the domain $[0,T] \times \Omega$, with $T > 0$. A mathematical copilot might successfully identify useful and new energy functionals by exploring various chosen multipliers and then attempting to derive useful quantities that can be brought under control by manipulating the resulting integrals. Newly discovered functionals which can be controlled may provide valuable new advances in mathematicians' understanding of solutions to the corresponding equation(s).
\end{itemize}

We now consider typical workflows in low-dimensional topology, which is an area of mathematics that has a unique flavor and uses tools from geometry, algebra, PDEs, group theory, combinatorics, and mathematical physics. Hence, some of these workflows are more challenging to convert to data. Low-dimensional topology is the study of $n$-manifolds (certain topological spaces that are locally homeomorphic to $\mathbb{R}^n$) of dimension $n \le 4$. A knot is a simple closed curve embedded in $\mathbb{R}^3$. Knot theory plays a fundamental role in low-dimensional topology as every 3- and 4-manifold can be represented by a framed link; i.e., a collection of knots labeled by integers, called a Kirby diagram \citep{Kirby:1978}. We refer the reader to \cite{Juhasz2023book} for more detail on low-dimensional topology and knot theory and explanations of the mathematical terms used below.

\begin{itemize}
\item A central type of question in this area is classification, which requires being able to show whether two objects (e.g., knots or manifolds) are equivalent. The first step is finding a suitable representation. In case of knots, this could be a projection to the plane, called a knot diagram, a closure of a braid (a number of parallel strands running around a central axis), or a grid diagram. Knot diagrams can be encoded numerically as PD, DT, or Gauss codes, and braids as braid words. Manifolds can be represented using Kirby diagrams, as triangulations, branched covers along links, or as geometric objects (e.g., hyperbolic 3-manifolds).

To show that two representations of the same knot or manifold are equivalent, one finds a sequence of certain moves connecting them, which can be Reidemeister moves \citep{Reidemeister:1927Abhandlungen} in case of knot diagrams, Markov moves for braids, or Kirby moves in case of Kirby diagrams. These search problems lend themselves to techniques such as reinforcement learning; see \citep{Gukov:2021Learning}. The workflow steps can be encoded by a sequence of representations of the object, such that consecutive representations are related by one of the standard moves. Software packages such as SnapPy \citep{SnapPy} can be used to check whether these moves are valid and can also list valid moves. We can hence represent the workflow as follows:
\begin{center}
    \emph{choose type of representation and set of moves $\rightarrow$ find representations of two mathematical objects $\rightarrow$ repeatedly apply moves to first object $\rightarrow$ representation of second object.}
\end{center}

\item To show two objects are inequivalent, one defines invariants, which are typically algebraic objects (numbers, polynomials, or groups) that are unchanged by the above moves. These moves often also capture important topological properties of these objects. The most classical invariants are homology groups and the fundamental group. The Alexander polynomial is a knot invariant derived from the fundamental group of the knot complement. Representations of the knot group give rise to the more modern twisted Alexander polynomials. A recent knot invariant rooted in representation theory is Khovanov homology \citep{Khovanov:2000categorification}. Floer homology \citep{OzsvathSzabo:2004HF} and gauge theory \citep{Witten:1994monopoles} give rise to highly sophisticated knot and 3- and 4-manifold invariants. These invariants often give lower bounds on hard-to-compute topological quantities such as the 3- or 4-genus of a knot. SnapPy is capable of computing many of these knot invariants when run in SageMath \citep{sagemath}. A typical workflow would thus be as follows:
\begin{center}
    \emph{pair of mathematical objects $\rightarrow$ choose suitable invariant $\rightarrow$ compute invariant for objects $\rightarrow$ show invariants are inequivalent.}
\end{center}

\item To define a Floer-theoretic invariant, we construct a chain complex whose boundary map counts some pseudo-holomorphic curves with certain Lagrangian boundary conditions in some symplectic manifold. To show these counts are finite, one has to prove transversality results for the moduli spaces using difficult methods from PDEs, then obtain a formula for the dimension of the moduli spaces and compactify the moduli spaces. The next step is showing the boundary map squares to zero, in which case we have a chain complex, or sometimes we end up with a more complicated algebraic structure, such as an $A^\infty$-module or a differential graded algebra. One often has to use special coefficient systems, such as a Novikov ring, and deal with bubbling phenomena. There are many choices that go into the construction of our chain complexes, so one has to prove independence of these choices up to chain homotopy equivalence, which often relies on continuation maps and pseudo-holomorphic polygon counts. One then extracts more tractable invariants from the chain homotopy type using algebra, such as taking homology. Different invariants are often related by spectral sequences, defined using filtrations of the chain complexes. A similar workflow exists for gauge-theoretic invariants. More sophisticated invariants can be obtained by exploiting certain symmetries, such as $\mathbb{Z}_2$ or $\text{Pin}(2)$. In another popular direction, one can sometimes construct a homotopy type from chain complexes; see \citep{LipshitzSarkar:2018}. Certain constructions from algebraic topology can then be invoked to provide additional algebraic structure, such as Steenrod operations. We can hence represent the high-level workflow for defining a Floer-theoretic topological invariant as follows:
\begin{center}
    \emph{associate a symplectic manifold and Lagrangian submanifolds to our topological object $\rightarrow$ decide what pseudo-holomorphic curves to count $\rightarrow$ prove transversality of the moduli spaces $\rightarrow$ find dimension formula for moduli spaces $\rightarrow$ compactify moduli spaces $\rightarrow$ choose coefficients and the right algebraic structure $\rightarrow$ extract invariant using algebra $\rightarrow$ prove independence of choices.}
\end{center}
It is apparent from the complexity of the workflow that this is very difficult to turn into data. As the number of workflow steps increase, the number of datapoints has to increase exponentially to cover all possible combinations between steps (although not all combinations may be possible in all cases, which may mitigate this issue).

\item After an invariant is defined, we study what topological information it captures and derive applications. Here, it is important to know what applications are mathematically relevant. This workflow is very different from many other areas of mathematics, where tools are developed to tackle specific open problems, and should be compared with Section~\ref{sec: motivated proof} on motivated proofs. This step requires sophisticated intuition and a global vision. One way to turn this into data is to define pairs
\begin{center}
    \emph{(invariant, topological application)},
\end{center}
which could be used by a mathematical copilot to recommend potential applications of a given type of invariant. We want to choose applications that are related to problems from a list of important open questions.

\item As many of the invariants are hard to compute, it is crucial to develop methods to compute them. For Floer-theoretic 3- and 4-manifold invariants, for example, experts use surgery formulas, spectral sequences, grading arguments, exact triangles, gluing formulas, or computations in explicit diagrams in simpler cases. The difficulty lies in the fact that one has to solve non-linear PDEs to obtain the pseudo-holomorphic curve counts contributing to the boundary maps in the chain complex, which is not algorithmic. Knot Floer homology now has a combinatorial definition and can be computed by SnapPy. It is often helpful to restrict attention to special classes of objects, such as to the class of alternating knots, 3-braid closures, etc., and perform computations or prove results for these. A schematic of this workflow is the following:
\begin{center}
    \emph{invariant $\rightarrow$ method of computation $\rightarrow$ class of objects to restrict to}.
\end{center}
Hence, a potential data point could look like
\begin{center}
    \emph{(invariant, method of computation, class of objects)}.
\end{center}

Certain knot invariants are easy to define, but no algorithm is known to compute them, such as the unknotting number or the 4-ball genus, and topologists use computable invariants, many of them arising from knot Floer homology, to give lower bounds on these. The paper \citep{davies2021advancing} describes a workflow for using supervised learning to find correlations between various invariants, which has led to an inequality relating the knot signature and hyperbolic invariants. Upper bounds can be obtained by performing certain moves on knot diagrams, possibly with the help of reinforcement learning or Bayesian optimization; see \citep{Gukov:2023SearchingRibbons} and \citep{unknotting2024}.
For example, one could take \emph{(4-ball genus, lower bound from Rasmussen $s$-invariant, torus knots)}, which leads to a solution of Milnor's conjecture on the 4-ball genus of torus knots \citep{Rasmussen:2010slice}.

\item The classification of smooth 4-manifolds is still wide open. There exist 4-manifolds that are homeomorphic but not diffeomorphic, which are called exotic pairs. To construct these, experts use methods from algebraic and symplectic geometry, such as blow-ups, fiber sums when given Lefschetz fibrations, and other gluings, typically along 3-tori, such as knot surgery. To show two 4-manifolds are homeomorphic, one shows they have isomorphic intersection forms and invokes Freedman's theorem \citep{Freedman:1982four-manifolds}, for which the fundamental group has to satisfy certain restrictions (e.g., being trivial). The intersection form and a presentation of the fundamental group can be read off a Kirby diagram. This fundamental group computation is often the most difficult step, and note that any finitely presented group can arise as the fundamental group of a closed 4-manifold. Furthermore, there is no algorithm to decide whether a finitely presented group is trivial. So, some questions in low-dimensional topology are beyond the reach of computers. As mentioned above, one can use Kirby calculus to show if two 4-manifolds are diffeomorphic. Swenton's Knot-Like Objects software\footnote{\url{https://community.middlebury.edu/~mathanimations/klo/}} is capable of performing Kirby moves. To show they are not diffeomorphic, one usually distinguishes them using the gauge-theoretic Seiberg--Witten invariants. No algorithm is known to compute these, but they can be computed in some instances using Taubes' non-vanishing result for symplectic 4-manifolds \citep{Taubes:1994}, together with various glueing results, such as Fintushel and Stern's knot surgery formula \citep{FintushelStern:1998}. A schematic of a typical such workflow in smooth 4-manifold topology is as follows:
\begin{center}
    \emph{pair of smooth 4-manifolds $\rightarrow$ find their Kirby diagrams $\rightarrow$ show they are simply-connected $\rightarrow$ show they have isomorphic intersection forms (hence homeomorphic by Freedman) $\rightarrow$ try to show they are diffeomorphic using Kirby calculus; if this fails $\rightarrow$ compute Seiberg--Witten invariants to show not diffeomorphic.}
\end{center}
The last step, the computation of Seiberg--Witten invariants, is discussed in more detail in the previous bullet point. Computer algebra systems can be useful for the computation of the fundamental group and the intersection form.
\end{itemize}

3-manifold topology has a completely different flavour. Here, methods from geometric group theory, hyperbolic geometry, and combinatorial topology dominate. These combinatorial methods include triangulations and normal surface theory. Geometric group theory studies groups using their actions on metric spaces (e.g., on the Cayley graph). Also note Perelman's proof of the 3-dimensional Poincar\'e conjecture using the Ricci flow, rooted purely in geometric analysis; see \citep{MorganTian:2007}. Well-developed computer packages exist to aid 3-manifold topologists, including the already mentioned SnapPy for hyperbolic 3-manifolds and Regina \citep{Burton:2004Regina} for triangulations. We do not provide specific workflows in 3-manifold topology due to the diversity of techniques used.

Having discussed differential equations and low dimensional topology, we now turn to the Langlands philosophy, a far-reaching series of conjectures relating the continuous to the discrete. The story starts with Gauss' law of quadratic reciprocity, a striking result explaining a rather counterintuitive relationship between two apparently unrelated questions; given two odd primes~$p$ and~$q$, there is an unexpected symmetry relating the questions of whether~$p$ is a square modulo~$q$, and whether~$q$ is a square modulo~$p$. Gauss' theorem can be thought of as one of the first nontrivial results in arithmetic. In the 1800s Eisenstein extended the result to higher powers, and ultimately the results were recast as class field theory, one of the highlights of 19th century mathematics. The advent of cohomology (a tool initially introduced to study questions in topology) after the war clarified things further, and by the 1950s the main proofs (now in the form of Artin reciprocity) were well-understood, although they even now remain frustratingly long.  For example the 1968 book by Cassels and Froehlich~\cite{cassels1976algebraic} which contains the proofs of the main theorems is over 300 pages long, and the proofs have not become substantially shorter since.

The bombshell dropped by Langlands in the 1960s was to observe that class field theory itself was merely the 1-dimensional case of a far more profound nonabelian reciprocity law relating objects from harmonic analysis (the continuous) and objects from arithmetic (the discrete). In short, there seemed to be a collection of analytic objects (``automorphic representations'') each of which was related to a corresponding algebraic object (a ``Galois representation''). Rest assured that the details of the (highly technical) definitions are irrelevant to the story being told here; the key point is that automorphic representations are objects coming from harmonic analysis, and Galois representations are objects coming from arithmetic, and these are (on the face of it) two very different areas of mathematics.

Whilst the initial idea was compelling, it took some time before the depths of Langlands' insights could be turned into rigorous mathematical \emph{statements}, with early attempts involving data whose existence was only conjectural (for example the global Langlands group of a number field, an object whose definition is still unclear). On the other hand, a wealth of evidence accrued very quickly, making it absolutely clear that beneath the ``philosophy'' there were some beautiful mathematical statements waiting to be proved. In the 60 years since Langlands' ideas were first enunciated, there have been generalizations to other domains of mathematics; there are local and global Langlands conjectures, a $p$-adic Langlands philosophy, a mod $p$ philosophy (both of which are still wide-open and whose statements are still not completely understood), and geometric Langlands conjectures, where our understanding is much better and there is even a recent announcement of a proof in many cases, which was \href{https://www.quantamagazine.org/monumental-proof-settles-geometric-langlands-conjecture-20240719/}{picked up by the media}, although the papers have not yet been published.

Let us stick to the classical local and global conjectures, which already raise plenty of questions. What do the workflows look like in this area? 

\begin{itemize}
  \item There is the fight to understand the question. Part of the subtlety of this part of mathematics is that whilst there is a general idea that certain automorphic representations (objects from analysis) should be in correspondence with certain Galois representations (objects from arithmetic), it is still not clear how to uniquely pin down such a correspondence (for general reductive groups over number fields, say). One can write down a list of criteria which the correspondence should satisfy, but what mathematicians really seek is some kind of constructive or ``canonical'' construction of an object on one side of the conjectured correspondence, given an object on the other side. Without this definition, it is impossible to formally state the theorems we would like to prove. In particular, the main goal here is not to compute a value or to prove a theorem, it is to construct a function.
  \item However there are plenty of instances of rigorous mathematical conjectures which we can write down and for which we have ample evidence in the form of both numerical examples and partial results. Here is a conjecture which can be made completely precise. Let $n$ be a positive integer. Given an algebraic automorphic representation for the group $GL(n)$ over a number field, one expects to be able to associate to it a compatible family of $n$-dimensional $p$-adic Galois representations of this number field, and conversely given the family of Galois representations one expects the existence of an automorphic representation. This statement is fully proved in the case $n=1$ by class field theory. What little we do know for $n>1$ is for the most part restricted to totally real and CM number fields (which together form 0 percent of all number fields, one fact indicating the depths of our ignorance) and what litle we do know has been very hard-won. The key difficulty is to find some kind of bridge taking us between the discrete and the continuous, and there are limited tools available. Certain automorphic representations can be studied using differential geometry. Serre's work linked differential geometry and algebraic geometry in the 1950s; Grothendieck's revolution of algebraic geometry in the 1960s gave links to arithmetic, and thus provided the first viable bridge in the case of $n=2$. Algebraic geometry was the source of the first general constructions of elliptic curves from modular forms, provided by Eichler and Shimura at the end of that decade. This work was simultaneously a milestone achievement and a miniscule special case of the Langlands program for $GL(2)$ over the rationals. We had to wait for 25 years before Wiles and Taylor managed to go from an elliptic curve back to a modular form; Fermat's Last Theorem was a consequence. In particular a major problem in number theory was resolved by proving an extremely small part of Langlands' conjectures. One of the major difficulties here is that the arguments need nontrivial inputs from arithmetic, algebra, geometry and analysis, and would take thousands of pages to explain if written out in full. These thousands of pages establish just a very small special case of the Langlands program for $GL(2)$ over the rationals. In some sense it does not bode well that the $GL(1)$ proofs were hundreds of pages long and even a small fragment of the $GL(2)$ proofs are thousands of pages. Within these pages there are of course several profound ideas; many more seem to be needed. There are few living humans who have read and completely understood the full proofs of the modularity theorem linking elliptic curves and modular forms; on the other hand it is certainly possible to understand a high-level overview, and most people take on parts of the argument as ``black boxes'' where the statements of the results are comprehensible but the proofs might involve techniques which are not part of their expertise. In particular the workflow here does not involve understanding all the details of everything; communication between mathematicians with different expertise is key here, as well as taking results from other areas of mathematics on trust.
  \item The success of Wiles' program showed in particular that algebraic geometry gave a viable route to cross the chasm between the discrete and the continuous and link objects on both sides. However the techniques will only apply when certain quite restrictive conditions hold. Another tool in the area is ``Langlands functoriality'', a tool for producing new analytic objects from old ones based on predictions on the arithmetic side of the picture. If the Langlands conjectures are true then basic constructions in arithmetic should have analogues on the analytic side, and one can independently try and make those constructions. Another motivation for making such constructions is that one might start with an analytic object for which the algebro-geometric tools do not apply, and then attempt to massage it until they do apply. Many of the results known for $GL(n)$ for $n>2$ are proved in this way; the group $GL(n)$ does ``not admit Shimura varieties'' for $n>2$, meaning that algebraic geometry has limited applicability; however if one can use analytic means to move from an automorphic representation on $GL(n)$ to an automorphic representation on a unitary group (where the methods do work) then this enables more analytic objects to ``cross the bridge'' to the algebraic side. Results in Langlands functoriality thus have the dual role of adding weight to the conjectures (by proving consequences of them) and also enabling us to prove more cases of the conjectures. Here the work is Fourier-analytic and group-theoretic, with less reliance on algebraic and arithmetic geometry.
  \item As is often the case in mathematics, people attempt to understand the questions in this area by generalizing them. The Galois representations which the Langlands philosophy are $p$-adic in nature, and can move freely in $p$-adic families. This raises the question of whether one can $p$-adically deform automorphic representations; initially this sounds ridiculous because these objects are inherently complex-analytic. However $p$-adic analytic variants of the theory have now been established, and this gives us new freedom, unlocking new tools to establish more cases of the classical program; the key observation here is that $p$-adic analysis is far more closely related to arithmetic than classical analysis. What is happening in this area is a mixture of insights gleaned from analogy (``what if there is a more arithmetic version of analysis?'') turn into an actual theory (the existence of $p$-adic automorphic representations) but then must be backed up by very technical work showing that the theory can actually be used to solve new cases of the problem at hand. Work in this area thus needs to blend heuristic ideas and technical calculations. 
\end{itemize}

\paragraph{Further considerations.} An important aspect that needs to be considered when mapping mathematical workflows to data is that different representations of the same data can lead to different mathematical properties that can change the  method of proof, as well as the foundational model most suited for the chosen representation. 

For example, it is an open problem in knot theory whether there exists a polynomial-time algorithm that can detect whether a given knot, which is an embedded circle in $\mathbb{R}^3$, can be simplified (without breaking it) to a standard round circle. Instead of describing the knot as a curve in $\mathbb{R}^3$, knots are often specified in terms of their projection to two dimensions, and data representations of the knots keep track of which strands are above and which are below along the projection ray. 

The above-mentioned question then becomes to find a sequence of deformations of the embedded circle such that its projection has no crossings. It is known~\citep{Kauffman:2012Hard} that for some data representations of the projected knot (such as Dowker–Thistlethwaite codes~\citep{Dowker:1982Classification} with Reidemeister moves as allowed deformations~\citep{Reidemeister:1927Abhandlungen}), there exist examples where the number of crossings needs to be increased before all crossings can be removed. In another representation (grid representation with Dynnikov moves as allowed deformations), the number of crossings is monotonically decreasing~\citep{Dynnikov:2006Monotonic}, but both the representation of the knot as a grid diagram and the carrying out of the simplification steps are more complicated. These distinct representations of the same mathematical object lend themselves to different foundational models~\citep{Gukov:2021Learning, Kauffman:2022Rectangular}; for example, braid word representations of knot projections are closer to natural language, whereas other representations are closer to vision tasks or graphs. The different representations also inform the type of algorithm that needs to be used. If the simplification is non-monotonic, a local search can get stuck in local minima.

One common activity in mathematics, when confronted with a new statement of unknown truth, a conjecture, is to engage in the search for proof -- or a counterexample. This comes with its own workflows. The search for counterexamples to conjectures is particularly useful for hard problems with a truth certificate that can be verified in polynomial time, such as NP-hard or NP-complete problems. The idea is to cast the search problem into a Markov Decision Problem (MDP) whose terminal states are counterexamples, and attempt to solve the MDP using data science techniques. In the past, deep reinforcement learning has proven to be a powerful tool. If the RL agent finds a solution to the MDP, its episodic rollouts provide truth certificates for the counterexample, thus establishing a verifiably correct proof by counter-example, see e.g.~\citep{Gukov:2024Rigor} for a recent summary of these ideas and~\citep{Gukov:2021Learning,Wagner:2021Constructing,Gukov:2023SearchingRibbons,Charton:2024Patternboost} for some recent concrete applications.

In the formal setting, Learning to Disprove frames counterexample search in Lean 4 as proposing a candidate counterexample and then certifying it by producing a Lean proof term, using a large synthetic corpus of counterexample instances~\citep{learningtodisprove2025}.

\subsection{Meta-Reasoning}
\label{subsec:metareasoning}

Many of the current evaluations of mathematical reasoning abilities focus in reasoning inside a particular problem. In a mathematics competition or a class, that is generally enough: it is assumed that all problems are solvable, that solutions are not too long, and furthermore that all the necessary tools come from a known, bounded set (e.g., of topics allowed in competitions, or pre-requisites for the class). At the research frontier, however, a substantial amount of work is spent at the meta-level, where these assumptions are constantly being evaluated. The failure to prove a statement might be due to it being false (and perhaps a counter-example can be extremely non-trivial to produce). Or, the appropriate tools might not yet exist. In either case, the problem-level assumptions above might be violated, and one has to constantly decide whether to keep pushing on the current plan, and what to do if not. To the best of our knowledge, none of the current evaluations for mathematical reasoning directly probe these meta-level judgements, even though they represent an important opportunity for assistance from research copilots.

Some of the most basic meta-judgements that we might want to evaluate are (a) is a given statement true? (and does that change in light of given failed attempts to prove it?) (b) if it is, how much work will it take to prove it (or, given a partial proof, how much work is left?), and (c) is a given strategy adequate to prove it? In the formal setting, LeanProgress trains a predictor of proof progress from Lean proof states and uses it to guide search on longer Mathlib proofs~\citep{leanprogress2025}.
Many examples that we could use to evaluate the ability to answer these questions could be extracted from finished works: papers have examples of true statements; removing assumptions from those statements often renders them false (and most papers justify why the assumptions are needed, if not apparent, such as by constructing counter-examples where those assumptions are violated). We can measure the complexity of existing proofs in various ways (e.g., length, or by their dependencies inside a paper) to evaluate predictions of difficulty --- these can also be made relative, e.g., identifying which lemma in the context of a broader proof is going to take the most work. On the other hand, many important meta-level decisions cannot be readily observed in finished work, such as when to change the problem locally (such as adding stronger assumptions), or switching problems entirely --- perhaps temporarily, e.g., to gain intuition on a special case first, or more firmly, and in that case deciding what to work on instead. For such decisions, one might need to observe mathematicians \emph{doing their work}, as opposed to their finished products, as we discuss next.

\subsection{Data Collection in Real Environments}
\label{subsec: data collection}

To gather representative data on mathematical workflow steps, as well as other metamathematical items, such as proof heuristics, limitations of certain proof techniques, etc., it may be desirable to observe and absorb the full process of producing mathematics, including all data that is not represented in written, textual form. For an example of real-time narration of thought processes that arise when doing competitive problem-solving, see co-author Tim Gowers' series of YouTube videos.\footnote{\url{https://www.youtube.com/@TimothyGowers0/videos}}

Unfortunately, many of these intermediate process steps become evident only implicitly from data sources. For example, at the elementary level, various workflows to solve various limits,
such as $\lim_{x \to 0} \frac{\sin x - x}{x^3}$ or $\lim_{x \to \infty} x^{1/x}$,
which require repeated applications of known theorems, such as de l'H\^{o}pital's theorem, or related ones to resolve indeterminate cases, are distilled only by solving a large number of exercises; no explicit annotations for these workflows exist. On a more advanced level, these are often conveyed in blogs, talks, and oral discussions between mathematicians at conferences -- and not in the typical data sources used to train LLMs.

In principle, one could implement longitudinal studies that track research projects from inception to completion. However, such data collection would require classifying, structuring, and recording different tasks throughout the mathematician's daily work. The apparent challenge is to scale such approaches while keeping the overhead as small as possible for the researcher whose work is tracked.
The KWARC group at FAU Erlangen-Nürnberg has developed infrastructure for semantic mathematical documents and tested such systems with hundreds of students per semester~\citep{kohlhase2006omdoc, berges2023learning}, demonstrating the feasibility of structured mathematical data collection at scale, though their focus has been primarily on finished documents rather than the discovery process.

While such approaches are welcome to be tested, as a first alternative step, we recommend transcribing from online sources such as lectures, panel discussions, seminars, vlogs, etc. VideoMathQA evaluates mathematical reasoning grounded in lecture-style videos, including long videos, with questions that require integrating visual and spoken information across time~\citep{videomathqa2025}. While such data only requires minimal need for postprocessing and is often already collected in the form of videos, it typically covers broader mathematical aspects than textbooks. However, while humans can frequently generalize from a single instance of a certain workflow, LLMs often require a dataset that contains sufficiently many examples. 
While data collection effort will elicit higher-quality mathematical data, the time lag between the time when new research discoveries are made and when these are represented in lectures can be significant. On the other hand, at venues such as conferences, a lot of information about mathematics is exchanged, but not recorded. While it would technically be possible to record a portion of the conversations that happen at a conference in a privacy-preserving manner, and this would be a great source of data, as one would consistently have data that is at the forefront of research (unlike the mentioned panel discussions, where there is often a time lag), it would require a paradigm shift from conference participants to accept such data collecting measures. We urge further conversations about privacy-preserving ways of curating richer datasets on mathematical workflows, that minimally impinge on mathematicians' naturalistic practices. %

\subsubsection{From Artifacts to Trajectories in Formal Proof Engineering}
One distinctive advantage of formal mathematics is that the development environment already produces fine-grained signals that can be turned into datasets, which are in turn used to improve future systems. These signals include elaborator diagnostics, proof states, tactic traces, resource usage (heartbeats), and even lint warnings about unstable or redundant proof steps.  This makes formal proof engineering a promising setting for process-level datasets.

One such dataset type is a \emph{repair trajectory corpus}: sequences of (i) an attempted proof edit, (ii) compiler/LSP diagnostics (type errors, missing imports, failing goals, timeouts), and (iii) the subsequent patch. Such data directly represents the compile--diagnose--fix loop that dominates real formalization workflows, and it can be used to train copilots that are useful beyond one-shot ``fill the \texttt{sorry}'' tasks.

A second such dataset type is a \emph{refactoring corpus}: pairs of proofs before/after refactoring steps that improve readability, reduce instance pollution, decrease resource usage, or align with library style.  For example, proof refactoring guidance emphasizes extracting helper lemmas, compiling after each extraction, and using fast diagnostic feedback loops that 
(i) test compilation after each extraction, and then
(ii) use LSP diagnostics to locate natural boundaries.
Such datasets would support copilots that can shepherd code through review, not merely produce a valid proof.

As a third example, stability and hygiene linting (e.g., flagging tactic instability or unused steps) can be recorded as structured metadata. This enables benchmarks that measure not just whether a proof checks, but whether it is stable across versions and robust under realistic resource constraints.

\subsection{Motivated Proofs}
\label{sec: motivated proof}

We do not doubt that there is substantial value in bespoke AI tools that can automate specific elements of a mathematician's workflow. However, if the goal is not only to provide correct proofs but also to enhance a user's mathematical \textit{understanding} of the results~\citep{zhang2023cogsciMath} and guide new discoveries, we need new tooling -- and we argue, new kinds of data. In particular, we argue for data which faithfully represents the process of \textit{proof discovery} instead of \textit{proof exposition}. Towards this goal, we introduce \emph{motivated proofs}~\citep{polya1949or,morris2019motivated}, which contain and make transparent more of the proof discovery process. We argue for their value as a standard for LLM evaluation. 
We then provide some examples and discuss some preliminary observations about current LLM's ability to construct and identify motivated proofs. Finally, we speculate on a path towards large-scale evaluation of a model's ability to produce motivated proofs.

Broadly speaking, a motivated proof is one that makes clear to the reader where each step comes from.  For example, many interesting proofs require one to find a mathematical object with certain properties. An unmotivated proof will simply specify the object and check that it has the desired properties, while a motivated proof will explain how to arrive at the object.

\subsubsection{Examples of motivated proofs}
We present two theorems here with examples of motivated proofs. Further examples can be found in Appendix \ref{sec: thms_motivated}.
\paragraph{Cantor's theorem}
\label{sec:cantor}
Cantor's theorem states that there is no surjection from a set $X$ to its power set $\mathbb P(X)$. The proof proceeds by letting $f:X\to\mathbb P(X)$ be a function and trying to find a subset of $X$ that is not in the image of $f$. From here, an unmotivated proof will simply exhibit a set, namely, $\{x\in X:x\notin f(x)\}$, and verify that it is not in the image of $f$, which turns out to be straightforward.

By contrast, a motivated proof will systematically search for the required subset of $X$. Not knowing which subset to take, we can treat the subset as an unknown, just as we do when solving an equation, and try to narrow down the possibilities. The most general subset of $X$ can be expressed as $\{x\in X:P(x)\}$ for some as yet unspecified property $P$. We now want to prove, for an arbitrary element $y$ of $X$, that $f(y)\ne\{x\in X:P(x)\}$. So we need either an element $x$ of $f(y)$ such that $\neg P(x)$ or an element $x$ of the complement of $f(y)$ such that $P(x)$. There are not many elements around, so trying $y$ is one of the first things to do, and then we find that we need either $y\in f(y)$ and $\neg P(y)$ or $y\notin f(y)$ and $P(y)$. And now the property $y\notin f(y)$ is forced on us as our choice of $P$, and we end up with the same set as before, but this time with its origin explained.

\paragraph{Nilpotent units}

Now we consider an early result from commutative algebra. Let $R$ be a commutative ring, and let $x \in R$ be nilpotent, then $(1+x)$ is a unit, meaning it has a multiplicative inverse. This is typically proved by naming an element $y = \sum_{k=0}^{r-1} (-1)^k\cdot x^k$ and verifying that this is a multiplicative inverse element by calculation. We regard this as unmotivated, since the discovery process of finding $y$ was left out.

For a motivated proof, we need to search for an inverse element. Not knowing which element to take, we parametrize the most generic element we can. Since the only known elements of $R$ are $1$ and $x$, the most generic element is an integer polynomial in $x$, $\sum_{k=0}^m a_kx^k$. For this to be a right inverse of $1+x$, we must have that $1 = (1+x)(\sum_{k=0}^m a_kx^k) = a_0 + \sum_{k=1}^m (a_k + a_{k-1})x^k + a_mx^{m+1}$. For this to hold, we must eliminate all coefficients of the polynomial besides the constant term, which should be $1$. This gives us that $a_0 = 1$, $a_k = a_{k-1}$ for $1 \leq k \leq m$ and $a_m = 0$. The first two equations give us that $a_k = (-1)^k$ for all $k$, but this contradicts the last equation.

But we also know that $x$ is nilpotent, so if we have $r$ such that $x^r = 0$, then all coefficients from $x^r$ onwards can be ignored. This solves our issue, as we can let $m = r - 1$ to remove the $a_m = 0$ condition, and we are left with $a_k = (-1)^k$, so $(1+x)$ is a unit with inverse $\sum_{k=0}^{r-1} (-1)^k\cdot x^k$.

\subsubsection{Motivated proofs as an evaluation metric}
The standard for a proof's acceptance into the mathematical literature has historically been correctness rather than a completely motivated account. Moreover, mathematicians have been incentivized to condense and refine their proofs for reasons such as page limits in journals, which has further widened the gap between \textit{proof discovery} and \textit{proof exposition}. To assess a model's capabilities in aiding proof discovery, we suggest that holding the proof to the standard of being motivated provides advantages over requiring only correctness. 

As argued in ~\citep{polya1949or,morris2019motivated}, a motivated proof is more informative to the reader, as it provides more insight into how the proof was discovered and how the reader might discover similar results themselves. More speculatively, a model capable of producing motivated proofs will likely generalize better to novel problems. The first reason for this conjecture is that most results in the literature are not fully motivated, and a model that is capable of recognising this is less likely to blindly recall its training data. Requiring proofs to be motivated also restricts the amount of brute-force search a model can perform, so performance depends more on reasoning ability and less on test-time computational power. A dedicated study to test whether models that generate motivated proofs indeed generalize better to novel problems could be worth pursuing, because it might reshape how we approach automated reasoning.

We will now briefly evaluate current LLMs' ability to generate motivated proofs, as well as to evaluate whether a given proof is motivated.

\paragraph{Generation of motivated proofs}
As a preliminary investigation, we prompted o1-preview and Llama 3.2-90B to produce motivated proofs of five mathematical results. Our goal was not to use best-in-class LLMs since they typically differ only by a few percentages on common benchmarks; in contrast, we wanted to obtain a preliminary assessment of whether LLMs have the ability to work with the concept of a motivated proof at all. The results on which we tested the abilities to generate motivated proofs were selected to be elementary but to require a slightly non-trivial construction. After defining motivated proofs, we provided an in-context example of a motivated proof of the other four results and asked the model to generate a new motivated proof. If the model failed to provide a motivated proof, we would prompt it once to fix it, pointing out the unmotivated step. Each model was asked twice per problem, first with only examples of motivated proofs in the context, following a traditional chain-of-thought approach, then again with both examples and non-examples, mimicking contrasting chain-of-thought prompting. The problems and associated proofs we used in the prompts can be found in Appendix~\ref{sec: thms_motivated}.

In most responses, the model showed some evidence of ``understanding" the concept of motivated proofs by avoiding unmotivated constructions. However, this often led to pseudo-motivations, where some justification was given before a construction, but it didn't provide any useful substance besides obscuring the unmotivated step (and often appealed to some irrelevant fallacies). In the example below, the unmotivated step of Cantor's theorem is relegated to the statement ``An important question arises: Does $x$ belong to its own image $f(x)$?".

\fbox{\begin{minipage}{46em}
\textbf{o1-preview (excerpt)}

Our goal is to find a subset of \( X \) that cannot possibly be the image of any element under \( f \). To do this, we'll consider how elements of \( X \) relate to their images under \( f \).
\\ \\
\textbf{Exploring the Relationship Between Elements and Their Images:}
\\ \\
For each element \( x \in X \), \( f(x) \) is a subset of \( X \). An important question arises: \textbf{Does \( x \) belong to its own image \( f(x) \)?}
\\ \\
This leads us to categorize the elements of \( X \) into two groups:
\begin{enumerate}
\item Elements that are \textbf{members} of their image: $x \in f(x)$.
\item Elements that are \textbf{not members} of their image: $x \notin f(x)$.
\end{enumerate}
\end{minipage}}

The other common issue was the model making completely unsound deductions when trying to motivate their construction, only to arrive at the correct result by chance. This behavior is consistent with the model not fully understanding the result but having seen it in their training data and attempting post-hoc rationalization. Below is an example:

\fbox{\begin{minipage}{46em}
\textbf{Llama-3.2 (excerpt)}

We want to find a subset $Y \subseteq X$ such that $\phi(x) = Y$ for some $x \in X$. However, we don't know which subset to choose.

Let's try to find a property that $Y$ should satisfy. Since $\phi(x) = Y$, we know that $x \in \phi(x) = Y$. But we also know that $x \notin \phi(x)$, because if $x \in \phi(x)$, then $\phi(x)$ would contain at least two elements: $x$ and $\phi(x)$.

This suggests that $Y$ should be a subset of $X$ that contains some elements of $X$, but not all of them. In particular, $Y$ should not contain $x$ itself.

With this in mind, let's define $Y = \{x \in X \mid x \notin \phi(x)\}$.
\end{minipage}}

Here, the contradictory claims that $x \in \phi(x)$ and $x \notin \phi(x)$ have no basis, and if true would imply an immediate contradiction, making the remainder of the proof redundant. Also, the condition for $Y$ is in fact the opposite of what is needed, but this error doesn't seem to affect the remainder of the model's reasoning.

Other problems provided more hopeful examples: a motivated proof for the large totients problem was successfully generated both times by o1-preview and once by Llama.

Overall, o1-preview seemed to outperform Llama on most problems. We also observed a slight improvement by using negative as well as positive examples to ``explain" the concept. As an unfortunate side effect, the extra information also increased the tendency to ``obscure" unmotivated arguments as discussed above.

\paragraph{Recognition of motivated proofs}
To conduct a larger scale investigation of a model's ability to generate motivated proofs, it would help if there was an existing ``judge" model which could determine whether or not a given proof was motivated. In any case, recognition of motivated proofs represents an easier sub-goal towards generating motivated proofs.

We performed a small, preliminary investigation of whether current state-of-the-art models can reliably distinguish a motivated proof from an unmotivated proof. We stress that this is only a preliminary investigation and do not claim to draw any strong conclusion, but hope that this will inspire the community to carry out larger evaluations in this regard and work towards a standard of evaluating motivated proofs.

We first tested the model's ability to identify whether a given proof was motivated. We tested the five problems from the previous section, where each problem had three proofs to be evaluated individually. This included two human-written proofs, which we judged to be paradigmatic examples of motivated and unmotivated proofs, one unmotivated machine-generated proof and, where available, one motivated machine-generated proof. The unmotivated machine-generated proof was the first correct but unmotivated proof generated by Llama-3.2 in the previous section. For the machine-generated motivated proof, we took the response in the previous section that we judged to be most motivated, which was generated by Llama in the integer sum problem and o1-preview for nilpotent units and large totients, while for the other problems no responses were deemed sufficiently motivated.

For some proofs where there were incorrect but largely inconsequential components, we manually altered some equations to enforce correctness and removed any opening or closing sentences that included the phrase ``motivated proof" to avoid misleading the model. As with generation, we prompted the model by explaining the definition of a motivated proof and giving an example and non-example for each of the other problems, where the order of examples and non-examples was alternated to ensure that the model wasn't learning the ordering. Each model was given three attempts per proof per problem, and the results are recorded below. We use HM, HU, MM and MU as abbreviations for \enquote{human motivated}, \enquote{human unmotivated} and \enquote{machine unmotivated} respectively.

\begin{center}
\begin{tabular}{| c | c c c c| c c c c| c c |}
    \hline
  & o1-preview & & & & Llama-3.2 & & & & Total & \\ 
 & HM & HU & MM & MU & HM & HU & MM & MU & TP & TN \\
 \hline
 Cantor's theorem & 3/3 & 3/3 & - & 3/3 & 3/3 & 0/3 & - & 0/3 & 6/6& 6/12\\  
 Small doubling & 3/3 & 3/3 & - & 2/3 & 3/3 & 3/3 & - & 2/3 & 6/6 & 10/12\\
 Integer sum & 3/3 & 3/3 & 3/3 & 2/3 & 2/3 & 3/3 & 3/3 & 0/3 & 11/12 & 8/12\\
 Nilpotent units & 3/3 & 3/3 & 3/3 & 3/3 & 2/3 & 3/3 & 3/3 & 0/3 & 11/12 & 6/12\\
 Large totients & 3/3 & 3/3 & 3/3 & 3/3 & 3/3 & 2/3 & 3/3 & 1/3 & 12/12 & 9/12\\
 \hline
 Total & 15/15 & 15/15 & 9/9 & 13/15 & 13/15 & 8/15 & 9/9 & 3/15 & 46/48 & 39/60\\
 \hline
\end{tabular}
\end{center}

Under this setup, o1-preview performed strongly, while Llama was heavily biased towards accepting a proof as motivated. A possible explanation for Llama's poor performance in the MU set is that the machine-generated unmotivated proofs, unlike the human-generated ones, were attempts at generating motivated proofs, so these proofs contained some superficial features such as proof length and ``chatty" language that the model had associated with motivated proofs. 

In an effort to address this issue, we re-framed the experiment as a binary choice task. This was largely the same as above, but instead of being given a single proof to judge, the models were given two proofs and were told that one was motivated and one was unmotivated, and to judge which was motivated. Again we recorded the success rate over three valid attempts per (unmotivated,motivated) pair, per problem. The results are shown below.

\begin{center}
\begin{tabular}{| c | c c | c c | c |}
\hline
  & o1-preview & & Llama-3.2 & &Total \\ 
  Unmotivated proof source & Human & Model & Human & Model & \\
 \hline
 Cantor's theorem & 3/3 & 3/3 & 3/3 & 3/3 & 12/12 \\
 Small doubling & 3/3 & 3/3 & 3/3 & 3/3 & 12/12 \\
 Integer sum & 6/6 & 6/6 & 6/6 & 6/6 & 24/24 \\
 Nilpotent inverses & 6/6 & 3/6 & 0/6 & 2/6 & 11/24\\
 Large totients & 6/6 & 4/6 & 6/6 & 5/6 & 21/24\\
 \hline
 Total & 24/24 & 19/24 & 18/24 & 19/24 & 80/96 \\
 \hline
\end{tabular}
\end{center}

This improved evaluation method narrowed the gap between the two models, especially when using the MU proof which Llama previously struggled with, providing some evidence that the binary choice strategy reduces the model's tendency to be misled by superficial features common to motivated proofs. 

We stress that this is only a preliminary investigation, and that a larger investigation would be required to make strong claims about an LLM's ability to judge motivated proofs. We welcome the community to conduct a larger investigation, advising the following notes of caution based off our experience and some speculation:
\begin{itemize}
    \item It is easier for models to decide between two proofs, which is motivated, than to make an absolute judgement of a single proof.
    \item Models can be sensitive to the order in which the proofs are presented, so it is important to permute these (including the in-context examples).
    \item Models can be misled by superficial features common to motivated proofs, so it is important that false examples also have these features. We achieved this to some extent by using an LLM's failed attempt at writing a motivated proof, but this could potentially also be achieved with care in a human-generated dataset.
\end{itemize}

\paragraph{Replication with newer models}

Given the significant improvements in model performance since the release of an earlier preprint, we have replicated these investigations with newer models. Although these are still preliminary investigations, we acknowledge the risk of data contamination.

We first replicated the generation task, prompting each model to generate a motivated proof of each result given an example and non-example of each other result. Each task was repeated three times, with each response manually classified by the authors and the results tabulated below:

\begin{center}
\begin{tabular}{| c | c c c c | c |}
\hline
&
\begin{tabular}{c}
OpenAI\\GPT-5.2 Pro
\end{tabular}
&
\begin{tabular}{c}
DeepSeek\\Prover-V2
\end{tabular}
&
\begin{tabular}{c}
Google\\Gemini~3~Pro
\end{tabular}
&
\begin{tabular}{c}
OpenAI\\GPT-OSS~120B
\end{tabular}
&
Total
\\
\hline
Cantor's theorem     & 3/3 & 0/3 & 3/3 & 0/3 & 6/12  \\
Small doubling       & 1/3 & 0/3 & 2/3 & 1/3 & 4/12  \\
Integer sum          & 3/3 & 0/3 & 3/3 & 3/3 & 9/12  \\
Nilpotent inverses   & 3/3 & 0/3 & 1/3 & 3/3 & 7/12  \\
Large totients       & 3/3 & 3/3 & 3/3 & 3/3 & 12/12 \\
\hline
Total
& 13/15
& 3/15
& 12/15
& 10/15
& 38/60 \\
\hline
\end{tabular}
\end{center}

This simple investigation was sufficient to observe substantial variation between models, although we believe that more difficult problems are needed to meaningfully distinguish the top models. It is also noteworthy that both closed-source models consistently succeeded on Cantor's Theorem, which neither of the older models could succeed on. It is also interesting that DeepSeek Prover-v2 performed poorly despite its mathematical specialization, which is consistent with the hypothesis that heavy fine-tuning for formal proof generation may reduce performance on stylistically different proofs.

We then repeated the binary classification task with newer models, concentrating on the more difficult and relevant machine-machine distinction, using proofs generated in the previous task. The first (motivated,unmotivated) pair was generated by GPT-5.2 Pro and DeepSeek Prover respectively, and the differences were stark enough that every model was correct 100\% of the time. For the second pair we compared Gemini's correct responses to a varied selection of incorrect model responses (including Gemini's where possible). We used a human-written unmotivated proof for the problem of ``large totients", since none of the newer models produced an unmotivated proof. The performance of each model on the second pair is shown below.

\begin{center}
\begin{tabular}{| c | c c c c | c |}
\hline
Theorem
&
\begin{tabular}{c}
OpenAI\\GPT-5.2 Pro
\end{tabular}
&
\begin{tabular}{c}
DeepSeek\\Prover-V2
\end{tabular}
&
\begin{tabular}{c}
Google\\Gemini~3~Pro
\end{tabular}
&
\begin{tabular}{c}
OpenAI\\GPT-OSS~120B
\end{tabular}
&
Total
\\
\hline
Cantor's theorem & 2/3 & 1/3 & 3/3 & 1/3 & 7/12 \\
Small doubling & 3/3 & 0/3 & 0/3 & 3/3 & 6/12 \\
Integer sum & 3/3 & 3/3 & 3/3 & 3/3 & 12/12 \\
Nilpotent inverses & 3/3 & 1/3 & 3/3 & 3/3 & 10/12 \\
Large totients & 3/3 & 3/3 & 3/3 & 3/3 & 12/12 \\
\hline
Total
& 14/15
& 8/15
& 12/15
& 13/15
& 47/60 \\
\hline
\end{tabular}
\end{center}

This generally shows some improvement in distinguishing motivated from unmotivated proofs, although some more subtle distinctions are still consistently missed.

\paragraph{Recommendations}

The existing mathematical literature, machine learning datasets and output evaluation standards are concentrated mostly on proof correctness, and our principal recommendation is to increase emphasis on proof motivation to better reflect the proof discovery process. One natural path forward would be to create a corpus of mathematical results with corresponding motivated and unmotivated proofs, and then to conduct a detailed evaluation of current models' ability to distinguish these. Once it is ascertained that models can judge motivation with high accuracy, one can design metrics for motivated proof generation.

\section{Conclusion}

We have outlined difficulties that pertain to natural language as well formal language mathematics datasets that we believe are present hindrances to the progress of AI systems towards becoming real \emph{mathematical thought partners} that are as, if not more, useful to mathematicians as GitHub's Copilot\footnote{\url{https://github.com/features/copilot}} is to programmers.

The advantages and disadvantages of natural language and formal language datasets are frequently complementary. What is easy in one representation of mathematics is often hard in the other. For example, automatic evaluation is easy in formal language but hard in natural language; representing rich interaction modes is often comparatively easy in natural language but harder to express in formal language. 

We have identified several facets of mathematical practice that are currently not represented in the data used to design and evaluate the deployment of AI systems for mathematics, such as various workflows. We acknowledge that the aspects of workflows we highlight here likely do not capture \textit{all} possible facets of mathematical practice - but we believe they represent an important start, offering an outline for what we may be able to curate to more human-compatible and explainable mathematical AI systems.%

While our aim has been exclusively in mathematics, these approaches of mapping processes of scientific discovery to data are not restricted to mathematics and may be adapted to other scientific domains as well. Some of the workflow items, like those related to literature search 
are directly relevant to other domains. We urge the community to explore how intermediate steps in the process of scientific discovery look like in other areas of science as well.

Regarding purely proof creation, time will tell whether human-annotated proofs are necessary--or whether a purely formal approach together with ingenious proof search techniques will succeed. The comparatively short history of machine learning has taught us to expect the unexpected. In the short term, however, it is imperative to establish better datasets and novel benchmarks for mathematics (either in natural language or formal language), to go beyond the current ones that only test proof or result creation - and also test the ability to assess different mathematical workflows, to summarize mathematics, to explain limitations of proof techniques etc. This will support the next generation of machine learning models and AI tools that can help us discover more mathematics fast and, subsequently assist with any other scientific discipline that uses mathematics as a foundation (e.g. systems biology that relies on ordinary differential equations, or physics that relies on several subfields of mathematics -- and inspires new subfields at the same time). At the same time, better mathematical copilots may have strong educational benefits. %
Recent controlled experiments suggest that LLM-based explanations can positively impact learning for high-school level mathematics, particularly when students attempt problems independently before consulting the model~\citep{Kumar2025MathEducationLLMs}. Moreover, Human-AI tutoring systems such as Tutor CoPilot have demonstrated measurable improvements in student outcomes when LLMs provide real-time pedagogical guidance to novice tutors~\citep{wang2024tutor}.

Having a deeper understanding of the processes by which one arrives at a proof (heuristics, workflows, etc.), which are all concentrated in the concept of a motivated proof, mathematical copilots can also teach the next generation of mathematical minds.

Even though the history of the concept of \enquote{proof} spans millennia, the story is ongoing, as currently, the search for a machine-learnable \enquote{proof data structure} is an important focus point and represents a new chapter in the ongoing story of what a proof really is.

\section*{Acknowledgements}
Thomas Lukasiewicz was supported by the AXA Research Fund. Katherine M. Collins acknowledges support from the Cambridge Trust. The work of Fabian Ruehle is supported by NSF grants PHY-2210333, PHY-2019786 (The NSF AI Institute for Artificial Intelligence and Fundamental Interactions), and startup funding from Northeastern University. Timothy Gowers would like to acknowledge generous support from the Astera Institute. We thank Terence Tao for useful remarks.

\clearpage

\bibliographystyle{unsrtnat}
\bibliography{references}
\clearpage
\appendix
\section{Appendix: Problems for Motivated Proof Experiments}
\label{sec: thms_motivated}
For the motivated proof experiments in \ref{sec: motivated proof}, we used five problems including the two examples. These are presented how they were presented to the models, including the human-written motivated and unmotivated proof used for in-context examples.
\paragraph{Cantor's theorem}
There is no surjection from a set $X$ to its power set $\mathbb P(X)$.

MOTIVATED PROOF: Let $f : X \rightarrow \mathbb P (X)$ be our function, we need to find some $S \subseteq X$ not in the image of $f$. Not knowing which subset of $X$ to take, we can treat the subset as an unknown, just as we do when solving an equation, and try to narrow down the possibilities. The most general subset of $X$ can be expressed as $\{x\in X:P(x)\}$ for some as yet unspecified property $P$. We now want to prove, for an arbitrary element $y$ of $X$, that $f(y)\ne\{x\in X:P(x)\}$. To obtain our contradiction, we need either an element $x$ of $f(y)$ such that $\neg P(x)$ or an element $x$ of the complement of $f(y)$ such that $P(x)$. There are not many elements around, so trying $y$ is one of the first things to do, and then we find that we need either $y\in f(y)$ and $\neg P(y)$ or $y\notin f(y)$ and $P(y)$. The choice $P(x) = x \notin f(x)$ satisfies this, so the set $S = \{x \in X : x \notin f(x)\}$ has the desired property.

UNMOTIVATED PROOF: Let $f : X \rightarrow \mathbb P (X)$ be our function, we claim that the set $S = \{x \in X : x \notin f(x)\}$ is not in the image of $f$, hence $f$ is not surjective. Suppose there exists $y \in X$ such that $f(y) = S$. But then $y \in f(y)$ iff $y \in S$ iff $y \notin f(y)$ by the definition of S, giving a contradiction.

\paragraph{Small doubling}
There exists a subset A of the natural numbers with cardinality n, such that the set $A + A = {a_1 + a_2 : a_i, a_j \in A}$ has cardinality $2n-1$.

MOTIVATED PROOF: We don't know which set to take for $A$ so we treat it as an unknown, just as we do when solving an equation, and try to narrow down the possibilities. Since we know that $|A| = n$, we can write $A = \{a_1,...,a_n\}$ for $a_1,...,a_n$ distinct, so we have $A+A = {a_i + a_j : i,j = 1,...,n}$. This isn't immediately helpful since we don't know how many collisions there will be in $A + A$, but we try to narrow down the possibilities. The simplest way to distinguish natural numbers is by ordering them, so we assume without loss of generality that $a_1 < ... < a_n$. Applying this monotonicity to our sums, we deduce that $a_i + a_j < a_i + a_k$ whenever $j < k$. 

Trying to distinguish as many elements as possible, we fix some $i$ and take this inequality to its logical conclusion, finding that $a_1 + a_i < .... < a_n + a_i$, or that we have n distinct elements. However, we can further extend this chain to the left if $1 < i$ or the right if $i < n$, giving us the longer chain $a_1 + a_1 < ... < a_1 + a_i < ... < a_n + a_1 < ... < a_n + a_n$, which has $2n-1$ distinct elements. If we want to have $|A+A| = 2n-1$, then these are all the elements of $A + A$, but we still have this free parameter $i$. Repeating the construction with some $j > i$ and observing the first syntactic different element, we notice that $a_1 + a_{i+1} = a_2 + a_i$, and isolating the variable i we notice that $a_2 - a_1 = a_{i+1} - a_i$. This defines an arithmetic progression with first element $a_1$ and common difference $a_2-a_1$, so let's see if that enough.

Suppose A is an arithmetic progression of length $n$, then $A = {a,a+d,\ldots,a+(n-1)d}$, then $A+A = {2a,2a+d,\ldots,2a+(2n-2)d}$ has size $2n-1$, completing the proof.

UNMOTIVATED PROOF:  Let $A = \{1,...,n\}$, which has cardinality n, then $A + A = \{2,...,2n\}$ has cardinality $2n-1$, completing the proof.

\paragraph{Integer sums}
There is a polynomial $P(x)$ with rational coefficients such that $P(n) = \sum_{k=0}^n k$ for all non-negative integers $n$.

MOTIVATED PROOF:  We can write a generic polynomial as $P(n) = \sum_{j=0}^m a_jn^j$. However, since $m$ is unknown the equation is reasonably unwieldy. Consequently, we wish to find an upper bound on the degree of $P(n)$. We do this by applying the trivial bound, $P(n) = \sum_{k=0}^n k \leq \sum_{k=0}^n n = n^2+n$, and since this equation holds for arbitrarily large $n$, we have that $m = \mathop{deg} P \leq 2$, so we can write $P(n) = a_0 + a_1n + a_2n^2$. To find $a_0$, $a_1$ and $a_2$, we can substitute some small values of $n$, so we have $a_0 = P(0) = 0$, $a_0 + a_1 + a_2 = P(1) = 1$ and $a_0 + 2a_1 + 4a_2 = P(2) = 3$. Solving the resulting linear equation, we have that $a_0 = 0$, $a_1 = a_2 = 1/2$ so $P(n) = \frac12(n^2+n)$.

To confirm that this works for all $n$, we can use induction. We have checked the base case already so we only need to do the induction step, so suppose that $\sum_{k=0}^n k = \frac12(n^2 + n)$, then we need to show that $\sum_{k=0}^{n+1} k = \frac12((n+1)^2 + (n+1))$. In order to use the induction assumption, we split the LHS to give us $\sum_{k=0}^n k + (n+1)$ and substitute in the induction assumption. By expanding and simplifying, we find that both sides of the equation are equal, completing our proof.

UNMOTIVATED PROOF:  We show that $P(n) = \frac12n(n+1)$ works using induction. First we observe that $P(0) = 0 = \sum_{k=0}^0 k$. Then suppose that $\sum_{k=0}^n k = \frac12n(n+1)$, then we have that $\sum_{k=0}^{n+1} k = \sum_{k=0}^n k + (n+1) = \frac12n(n+1) + (n+1) = (n+1)(\frac{n}2 + 1) = \frac12(n+1)(n+2)$, completing the proof.

\paragraph{Nilpotent units}

Let $R$ be a commutative ring, and let $x \in R$ be nilpotent. Then $(1+x)$ is a unit.

MOTIVATED PROOF: To show that $1+x$ is a unit, we need to find an inverse element. Not knowing which element to take, we parametrize the most generic element we can. Since the only known elements of $R$ are $1$ and $x$, the most generic element is an integer polynomial in $x$, $\sum_{k=0}^m a_kx^k$. For this to be a right inverse of $1+x$, we must have that $1 = (1+x)(\sum_{k=0}^m a_kx^k) = a_0 + \sum_{k=1}^m (a_k + a_{k-1})x^k + a_mx^{m+1}$. For this to hold, we must eliminate all coefficients of the polynomial besides the constant term, which should be $1$. This gives us that $a_0 = 1$, $a_k = a_{k-1}$ for $1 \leq k \leq m$ and $a_m = 0$. The first two equations give us that $a_k = (-1)^k$ for all $k$, but this contradicts the last equation.

But we also know that $x$ is nilpotent, so if we have $r$ such that $x^r = 0$, then all coefficients from $x^r$ onwards can be ignored. This solves our issue, as we can let $m = r - 1$ to remove the $a_m = 0$ condition, and we are left with $a_k = (-1)^k$, so $(1+x)$ is a unit with inverse $\sum_{k=0}^{r-1} (-1)^k\cdot x^k$.

UNMOTIVATED PROOF:  Let $r$ be such that $x^r = 0$ from nilpotency, and observe that $(1+x)(\sum_{k=0}^{r-1} (-1)^k\cdot x^k) = 1 + -x^r = 1$ by the formula for summing geometric progressions, hence $\sum_{k=0}^{r-1} (-1)^k\cdot x^k$ is an inverse of $1+x$ and $1+x$ is a unit.

\paragraph{Large totients}
We define Euler’s totient function $\phi$ as follows. If $n = \prod_{i=1}^m p_i^{k_i}$ is its prime factorisation, then $\phi(n) = \prod_{i=1}^m p_i^{k_i-1}(p_i-1)$. Show that for all $\epsilon > 0$, $\phi(n)/n$ can take values in $(1 - \epsilon,1)$.

MOTIVATED PROOF: We first try to simplify $\phi(n)/n$. Expressing $n$ in its prime factorisation to match the definition of $\phi(n)$, we have $\phi(n)/n = \prod_{i=1}^m (p_i^{k_i-1}(p_i-1)/p_i^{k_i}) = \prod_{i=1}^m (1 - 1/p_i)$. We notice that each factor is between $0$ and $1$, so the product is large where there are few factors so we let $m=1$ (or equivalently, letting $n = p^k$), giving us $\phi(n)/n = (1-1/p)$. Since this is clearly less than $1$, we only need to choose $p$ such that $1 - \epsilon < 1 - 1/p$, which simplifies easily to $p> 1/\epsilon$. Since there are infinitely many prime numbers, we can always choose such a $p$.

UNMOTIVATED PROOF:  Let $p$ be the smallest prime number such that $p > 1/\epsilon$, which exists as there are infinitely many primes. Then we have $\phi(p)/p = 1 - 1/p$, and as $p > 1/\epsilon$, we see that $1 - \epsilon < 1 - 1/p < 1$ as required.

\end{document}